%% file: final_version.tex
\PassOptionsToPackage{numbers}{natbib}
\documentclass{article}

\usepackage[preprint]{neurips_2025}

\usepackage[utf8]{inputenc} 
\usepackage[T1]{fontenc}    
\usepackage{hyperref}       

\makeatletter
\providecommand{\@trackname}{}
\makeatother

\usepackage{url}   
\usepackage{siunitx} 
\usepackage{booktabs}       
\usepackage{amsfonts}       
\usepackage{nicefrac}       
\usepackage{microtype}      
\usepackage{xcolor}         

\usepackage{tabularx}
\usepackage{multirow}
\usepackage{graphicx,verbatim}
\usepackage{amsmath, amssymb, bm}
\usepackage{subcaption}
\usepackage[table]{xcolor}

\newcommand{\dataset}{ESCAPE}

\title{A Standardized Benchmark for Multilabel Antimicrobial Peptide Classification}

\author{%
\textbf{Sebastian Ojeda}, \quad \textbf{Rafael Velasquez}, \quad \textbf{Nicolás Aparicio}, \quad \textbf{Juanita Puentes}, \\
\textbf{Paula Cárdenas}, \quad \textbf{Nicolás Andrade}, \quad \textbf{Gabriel González}, \quad \textbf{Sergio Rincón}, \\
\textbf{Carolina Muñoz-Camargo}, \quad \textbf{Pablo Arbeláez} \\
\\
Universidad de los Andes, Colombia}

\begin{document}

\maketitle

\begin{abstract}

Antimicrobial peptides have emerged as promising molecules to combat antimicrobial resistance. However, fragmented datasets, inconsistent annotations, and the lack of standardized benchmarks hinder computational approaches and slow down the discovery of new candidates. To address these challenges, we present the Expanded Standardized Collection for Antimicrobial Peptide Evaluation (\textbf{\dataset{}}), an experimental framework integrating over \num{80000} peptides from \num{27} validated repositories. Our dataset separates antimicrobial peptides from negative sequences and incorporates their functional annotations into a biologically coherent multilabel hierarchy, capturing activities across antibacterial, antifungal, antiviral, and antiparasitic classes. Building on \dataset{}, we propose a transformer-based model that leverages sequence and structural information to predict multiple functional activities of peptides. Our method achieves up to a \num{2.56}\% relative average improvement in mean Average Precision over the second-best method adapted for this task, establishing a new state-of-the-art multilabel peptide classification. \dataset{} provides a comprehensive and reproducible evaluation framework to advance AI-driven antimicrobial peptide research. 
\end{abstract}

\section{Introduction}
Antibiotics are crucial in modern medicine, enabling routine procedures and treating common infections. However, widespread misuse and overuse have led to the rise of antimicrobial resistance (AMR), where bacteria and other pathogens develop mechanisms to endure these drugs \cite{whoAntimicrobialResistance,mittal2020antimicrobials}. This issue extends beyond bacteria, including viruses, fungi, and parasites that have rapidly evolved, hindering the treatment of infectious diseases globally \cite{chambial2025frontiers}. The Institute of Health Metrics and Evaluation estimates that antimicrobial-resistant infections could cause over \num{39} million deaths between 2025 and 2050, with South Asia and Latin America facing the highest mortality rates \cite{healthdataGlobalBurden}. Besides the impact of AMR on healthcare systems and population lifespan, it is also a concern for national economies across the globe \cite{elsner2025global}. A recent United Nations report warns that, without action on AMR, not only will healthcare costs increase, but also the global GDP may decrease by US\$\num{3.4} trillion and drive an additional \num{24} million people into extreme poverty by 2030 \cite{unep2023amr}.

As a result, the scientific community has turned to alternative molecules capable of fighting infectious microorganisms without quickly triggering resistance. This process has led to the exploration of antimicrobial peptides (AMPs), which are either naturally occurring or synthetically designed proteins with a broad spectrum of antimicrobial properties \cite{xuan2023antimicrobial}. Unlike traditional antibiotics, AMPs often act through mechanisms harder for pathogens to evade, reducing the likelihood of resistance development \cite{islam2024antimicrobial}. Despite their therapeutic potential, the research and development of antimicrobial peptides, as with many pharmaceutical compounds, remains costly, time-consuming, and frequently unsuccessful. These difficulties are due to the inherent complexity of identifying effective AMPs and the extensive clinical trials and requirements novel drugs must undergo to reach the market and become profitable \cite{hamad2019superbugs}. These challenges create a bottleneck that limits the widespread adoption of AMPs and highlights the need for more efficient and scalable discovery protocols.

\begin{figure}[t]
    \centering
    \includegraphics[width=1\textwidth]{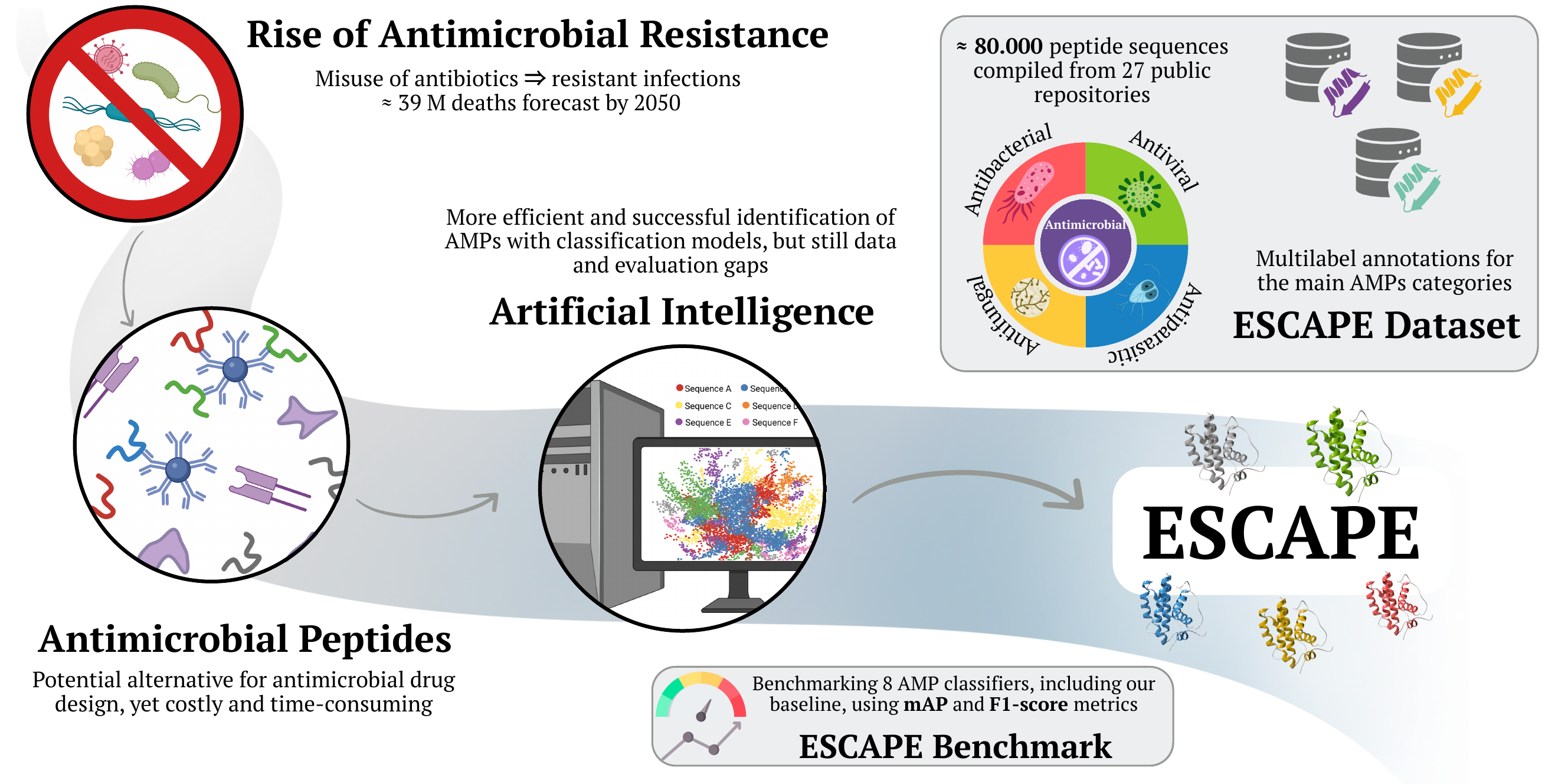}
    \caption{\textbf{Timeline of AMP Discovery and Computational Advances.} The rise of AMR underscores the urgent need for alternative therapies such as AMPs. While AI has shown promise in accelerating AMP discovery, progress is hindered by heterogeneous data and the absence of standardized evaluation protocols. We introduce \dataset{} to address these challenges and provide a robust foundation for future AI-driven methods.}
    \label{fig:overview-SCAPE}
\end{figure}

To address the challenges of AMP discovery, researchers have explored using Artificial Intelligence (AI) tools to accelerate the identification of promising antimicrobial candidates \cite{szymczak2023artificial}. Most existing models aim to predict whether a given peptide exhibits antimicrobial activity, often casting the task as a binary classification problem \cite{li2022amplify,lee2023amp,lawrence2021ampeppy}. While this binary approach helps identify potentially active peptides, it disregards the proven ability of AMPs to interact with multiple types of microorganisms \cite{de2023challenges}. Despite this constraint, interest in AI-driven AMP classification tools has expanded in recent years. Some of the proposed methods include models based on Graph Neural Networks (GNNs) and other architectures inspired by advances in Natural Language Processing (NLP) \cite{ruiz2022rational, orsi2024can}. However, the performance of these models remains suboptimal, indicating a need for more effective modeling strategies and further exploration of architectures.

A significant limitation in the current literature is the inconsistency and insufficiency of data selected for training AI models \cite{Yan2022AMPReview}. Although the number of publicly available datasets continues to grow, most AI methods rely on individual datasets containing only a few hundred to a few thousand peptides \cite{bty179}. Having such a limited number of examples from a single dataset when developing an AI model may restrict its learning potential and demonstrates the need for a more extensive set of data \cite{wang2022machine}. Furthermore, the absence of a standardized dataset or benchmarking framework makes it difficult to compare models reliably and to confidently identify which approach represents the state-of-the-art.

To overcome these limitations, we present three main contributions displayed in Fig.~\ref{fig:overview-SCAPE}. First, we compile, curate, and standardize \num{27} public AMPs databases into the Expanded Standardized Collection for Antimicrobial Peptide Evaluation (\dataset{}), a comprehensive dataset containing over \num{80000} peptides. We pre-process and validate all sequences to support robust AI-driven antimicrobial research. Second, we evaluate seven publicly available AMP classification methods on \dataset{}, adapting those designed initially for binary tasks to handle multilabel classification. To the best of our knowledge, this work results in the first benchmark that fairly compares existing approaches on a unified and scalable dataset for AMP multilabel classification. Finally,  we introduce the \dataset{} Baseline, a transformer-based architecture that leverages both sequence and structural information from peptides to predict not only whether it is antimicrobial, but also the types of pathogens it targets. Our baseline outperforms the state-of-the-art methods on the complete \dataset{} Dataset with a relative average improvement with respect to the second-best method of \num{2.56}\% and \num{1.90}\% in mean Average Precision (mAP) and F1-score, respectively.

\section{Related Work}

Given the global concern for antimicrobial resistance, there has been a noticeable rise in the development of AMP databases in recent years. Building on these resources, researchers have developed numerous AI models to analyze and identify potential AMP candidates to optimize antimicrobial drug discovery.

\subsection{AMP Databases}

In the search for novel AMPs, researchers have developed several databases, each containing peptides annotated with diverse biological activities \cite{claros2025computational}.
These databases can be broadly categorized into general \cite{yao2025dbamp,ma2025dramp,zhao2013lamp, singh2016satpdb} and specialized databases \cite{piotto2012yadamp,novkovic2012dadp,tyagi2015cancerppd}. General databases span a wide range of peptide functions or classes. In contrast, specialized databases focus on narrower aspects, such as peptide origin \cite{novkovic2012dadp} or the type of target organism 
\cite{piotto2012yadamp}, \cite{tyagi2015cancerppd}. The number of peptide entries and the granularity of functional classes vary substantially among databases. For instance, dbAMP \cite{yao2025dbamp} comprises \num{33065} peptides annotated with \num{58} distinct functional classes, while DRAMP \cite{ma2025dramp} includes \num{30260} entries but only \num{8} classes. Moreover, the hierarchical level of these classes is often inconsistent across databases. Namely, LAMP2 \cite{zhao2013lamp} annotates \num{23253} peptides into \num{38} classes, including “anti-Gram negative” and “anti-Gram positive,” which researchers can interpret as subclasses of a broader “Antibacterial” category.  In contrast, SATPdb \cite{singh2016satpdb} assigns its \num{19192} peptides to just \num{10} classes, among them a single “Antibacterial” label. Hence, currently available AMP datasets present key limitations. The class granularity and hierarchy discrepancies complicate training predictive models and comparing their performance across datasets. 

Additionally, training models on datasets composed entirely of peptides obtained using a single experimental technology may introduce methodological biases, ultimately hampering the model's ability to generalize across broader peptide sources \cite{al2024accelerating}. The \dataset{} Database addresses these limitations by integrating a wide range of public AMP datasets with more than \num{80000} peptides obtained from various sources. In addition, we standardize the class annotation system by curating a concise and biologically meaningful hierarchy of antimicrobial functions, thus improving the interpretability of annotations and facilitating dataset integration.

\subsection{AMP Benchmarks}

Benchmarking is critical in evaluating AMP prediction models, yet standardized protocols for consistent comparison are still missing. Many studies rely on custom datasets and train-test splits without releasing exact partitions, hindering reproducibility and fair comparison against other methods \cite{sidorczuk2022impact}. Furthermore, most benchmarking efforts focus on binary classification tasks that fail to capture the functional diversity and therapeutic relevance of AMPs \cite{zhang2021antimicrobial}. Although several studies have introduced multilabel classification approaches for AMPs \cite{ruiz2022rational,pang2022integrating}, the field still lacks a standardized, openly accessible benchmark designed to support rigorous evaluation in multilabel antimicrobial peptide prediction. In this context, the ESCAPE Benchmark is a significant advancement by enabling rigorous evaluation of seven state-of-the-art models on the multilabel AMP classification task, thereby facilitating fair and transparent comparisons of AI methodologies on a large-scale peptide dataset.

\subsection{AMP Classification Models}

There are two main categories of AMP classification methods in the current literature: those that rely exclusively on raw amino acid sequences and those that incorporate bioinformatically derived descriptors using specialized libraries. Sequence-focused methods include AMPlify \cite{li2022amplify}, which employs a bidirectional LSTM with multi-head and context attention to generate a summary vector. Other methods use Large Language Models (LLMs) \cite{LLMs} as the backbone of the architecture. For example, TransImbAMP \cite{pang2022integrating} uses a BERT model pretrained in a self-supervised manner via masked token prediction and fine-tunes a fully connected layer attached to its outputs for the AMP classification task. AMP-BERT \cite{lee2023amp}, another sequence-based model, also employs a pretrained BERT, but with an inserted class token whose embedding guides the classifier. More recently, dsAMPGAN \cite{zhao2024dsamp} integrates CNN, Attention, and BiLSTM layers with transfer learning to perform AMP classification and function prediction, while AMPpred-DLFF \cite{chen2024amppreddlff} combines ESM-2 \cite{esm-2} embeddings with graph attention networks and CNN modules to capture both spatial and sequential information.

In contrast, feature-augmented approaches compute additional descriptors before modeling. For instance, amPEPpy \cite{lawrence2021ampeppy} feeds CTD features (composition, transition, distribution of physicochemical amino-acid classes) to a Random Forest, and AMPs-Net \cite{ruiz2022rational} represents peptides as graphs enriched with physicochemical node and edge attributes to then process the peptides with a GNN. PEP-Net \cite{han2024pepnet} fuses one-hot amino-acid identities, computed physicochemical properties, and high-dimensional protein language model embeddings through residual convolutional and Transformer blocks to capture local information and global contextual information. AVP-IFT \cite{avp2024} employs a dual-branch framework integrating contrastive learning with a transformer network enhanced with biophysical and chemical properties. Recent work has also introduced ensemble-based feature-augmented approaches, including StackAMP \cite{karim2024stackamp}, AMP-RNNpro \cite{shaon2024amprnnpro}, and StackPIP \cite{yao2025stackpip}, which combine diverse peptide descriptors with multiple machine learning models to improve classification performance. Unlike prior methods, the ESCAPE Baseline introduces a bidirectional cross-attention mechanism by integrating the peptide sequence and its 3D distance representation, enabling the combination of spatial structural information with the sequence and leading to superior performance.

\section{Expanded Standardized Collection for Antimicrobial Peptide Evaluation}\label{Escape_Dataset_section}

Current research on AMP prediction faces a critical bottleneck due to fragmented, inconsistent, and small-scale datasets that vary widely in format, annotation standards, and functional coverage \cite{Yan2022AMPReview}. These limitations hinder the development of robust predictive models and complicate fair comparisons across methods. To overcome this gap, we introduce the \dataset{} Dataset, a unified collection of antimicrobial peptides compiled from \num{27} public repositories. This multilabel framework standardizes functional annotations to reflect the biological taxonomy of infectious agents, resulting in five classes: four main AMP activities (antibacterial, antifungal, antiviral and antiparasitic) and a fifth category (antimicrobial) that represents the general ability to act on any microorganism. Peptides that do not exhibit antimicrobial properties, and thus do not belong to any of the five classes, are considered Non-AMPs.

\subsection{Data Compilation} \label{data_compilation}

To build the \dataset{} Dataset, we collect experimentally validated and manually curated antimicrobial peptide entries from \num{27} public databases: BIOPEP-UWM Database \cite{minkiewicz2019biopep}, CPPsite 2.0 \cite{agrawal2016cppsite}, CAMPR3 \cite{waghu2016campr3}, TumorHoPe \cite{kapoor2012tumorhope}, APD3 \cite{wang2016apd3}, SPdb \cite{choo2005spdb}, ParaPep \cite{pretzel2013antiparasitic}, CancerPPD \cite{tyagi2015cancerppd}, BrainPreps \cite{van2012brainpeps}, Quorumpeps \cite{wynendaele2013quorumpeps}, YADAMP \cite{piotto2012yadamp}, LAMP2 \cite{zhao2013lamp}, Milkampdb \cite{theolier2014milkamp}, DADP \cite{novkovic2012dadp}, AntiTbPdb \cite{usmani2018antitbpdb}, PeptideDB \cite{PeptideDB}, NeuroPrep \cite{wang2015neuropep}, SATPdb \cite{singh2016satpdb}, BioDADPep \cite{roy2019biodadpep}, NeuroPedia \cite{kim2011neuropedia}, DFBP \cite{qin2022dfbp}, dbAMP 3.0 \cite{yao2025dbamp}, DRAMP 4.0 \cite{ma2025dramp}, AVPdb \cite{qureshi2014avpdb}, Hemolytik \cite{gautam2014hemolytik}, DBAASP v3 \cite{pirtskhalava2021dbaasp}, and UniProt \cite{uniprot2019uniprot}. Each source contributes unique peptide profiles across four functional classes: antibacterial, antifungal, antiviral, and antiparasitic. This structure encapsulates key differences in mechanisms of action, such as disarrangement of bacterial membranes~\cite{brogden2005antimicrobial}, inhibition of cell wall biosynthesis~\cite{mahlapuu2016antimicrobial}, and interference with viral assembly~\cite{qureshi2014avpdb}, among others.

For the integration of Non-AMP samples, we follow the methodology outlined in TransImbAMP \cite{pang2022integrating}, focusing on selecting non-antimicrobial sequences from UniProt \cite{uniprot2019uniprot}. We apply stringent exclusion criteria, removing entries associated with keywords such as “membrane,” “toxic,” “secretory,” “defensive,” “antibiotic,” “anticancer,” “antiviral,” or “antifungal”. To expand this set, we incorporate peptides from curated datasets known for their non-antimicrobial functions \cite{roy2019biodadpep,pretzel2013antiparasitic}. This dual strategy of exclusion-based filtering and targeted selection constructs a high-confidence negative set that effectively distinguishes non-AMP sequences, providing a robust contrast for supervised training. Supplementary Material Section A offers specific details on the creation of the dataset, regarding handling the compiled databases and their associated licenses.

\begin{figure}[t!]
  \centering
  \newsavebox{\bigimage}
  \newlength{\bigheight}
  
  \sbox{\bigimage}{%
    \begin{subfigure}[b]{.6\textwidth}
      \centering
      \includegraphics[width=\textwidth]{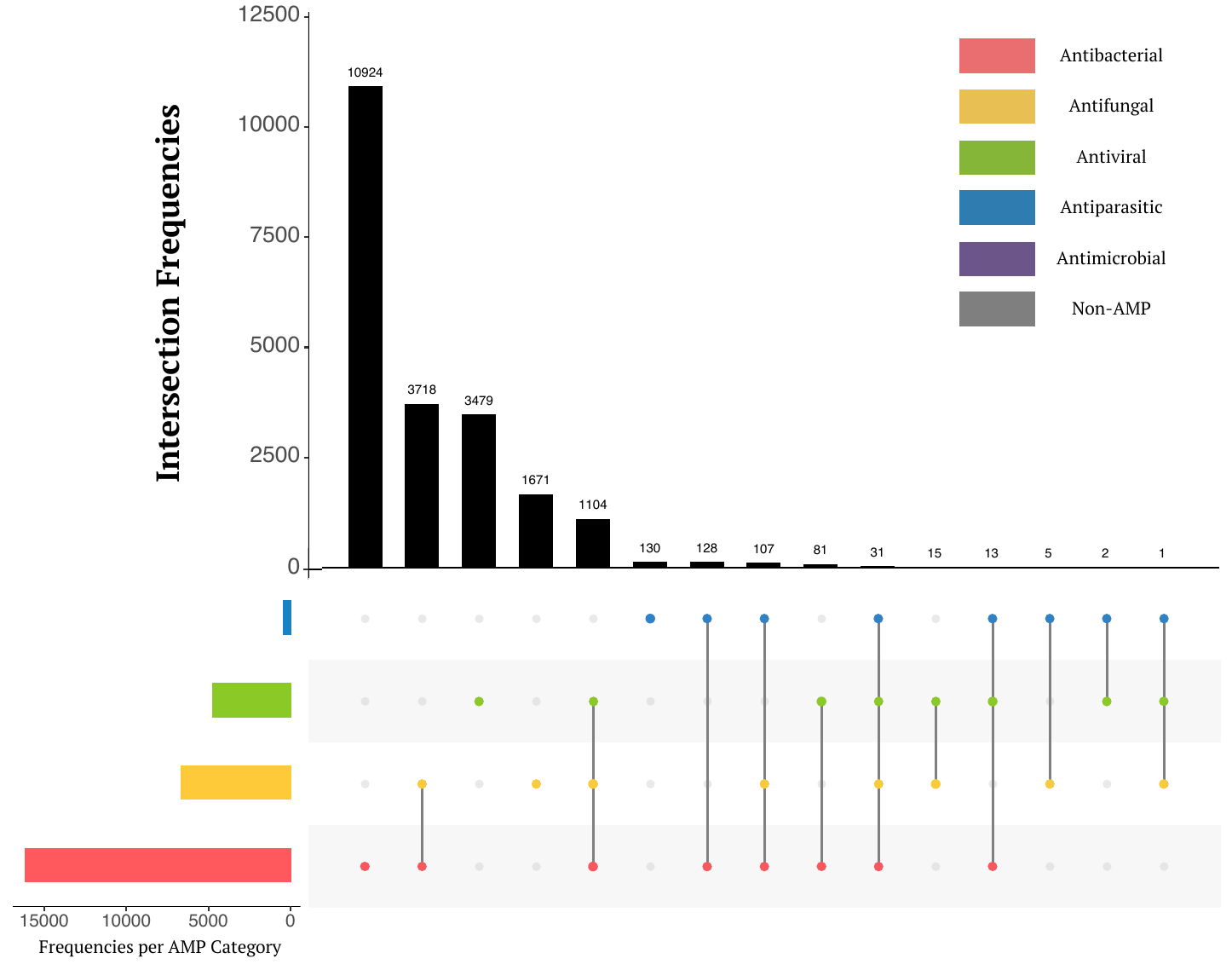}
      \caption{}
      \label{fig:upset-plot}
      \vspace{0pt}
    \end{subfigure}%
  }
  \usebox{\bigimage}%
  \hfill
  \begin{minipage}[b][\ht\bigimage][s]{.4\textwidth}
    \begin{subfigure}[t]{\textwidth}
      \centering
      \includegraphics[height=0.4\textwidth,width=\textwidth,keepaspectratio]{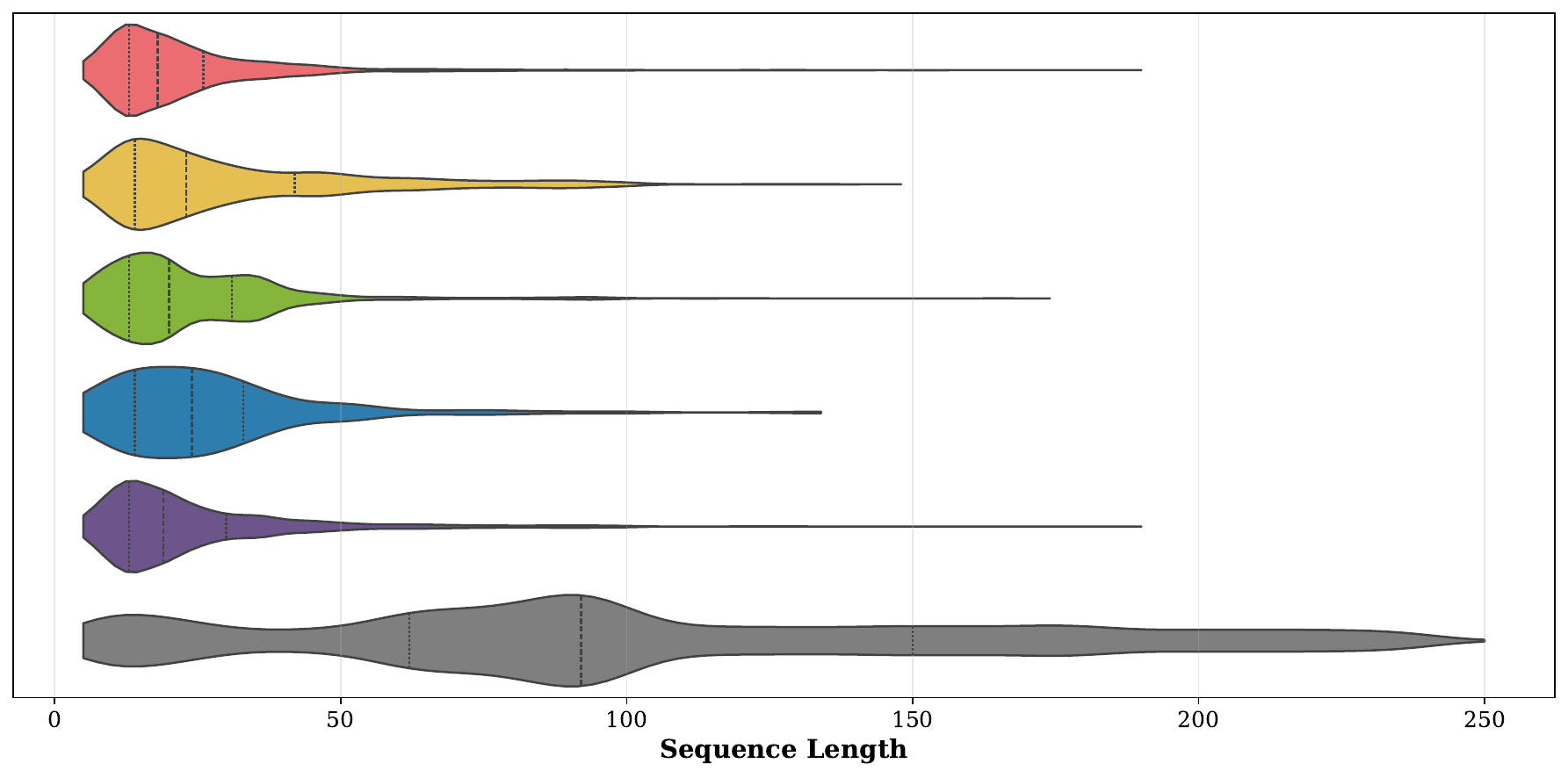}
      \caption{}
      \label{fig:violin-plot}
    \end{subfigure}


    \vfill

    \begin{subfigure}[t]{\textwidth}
      \centering
      \includegraphics[height=0.4\textwidth,width=\textwidth,keepaspectratio]{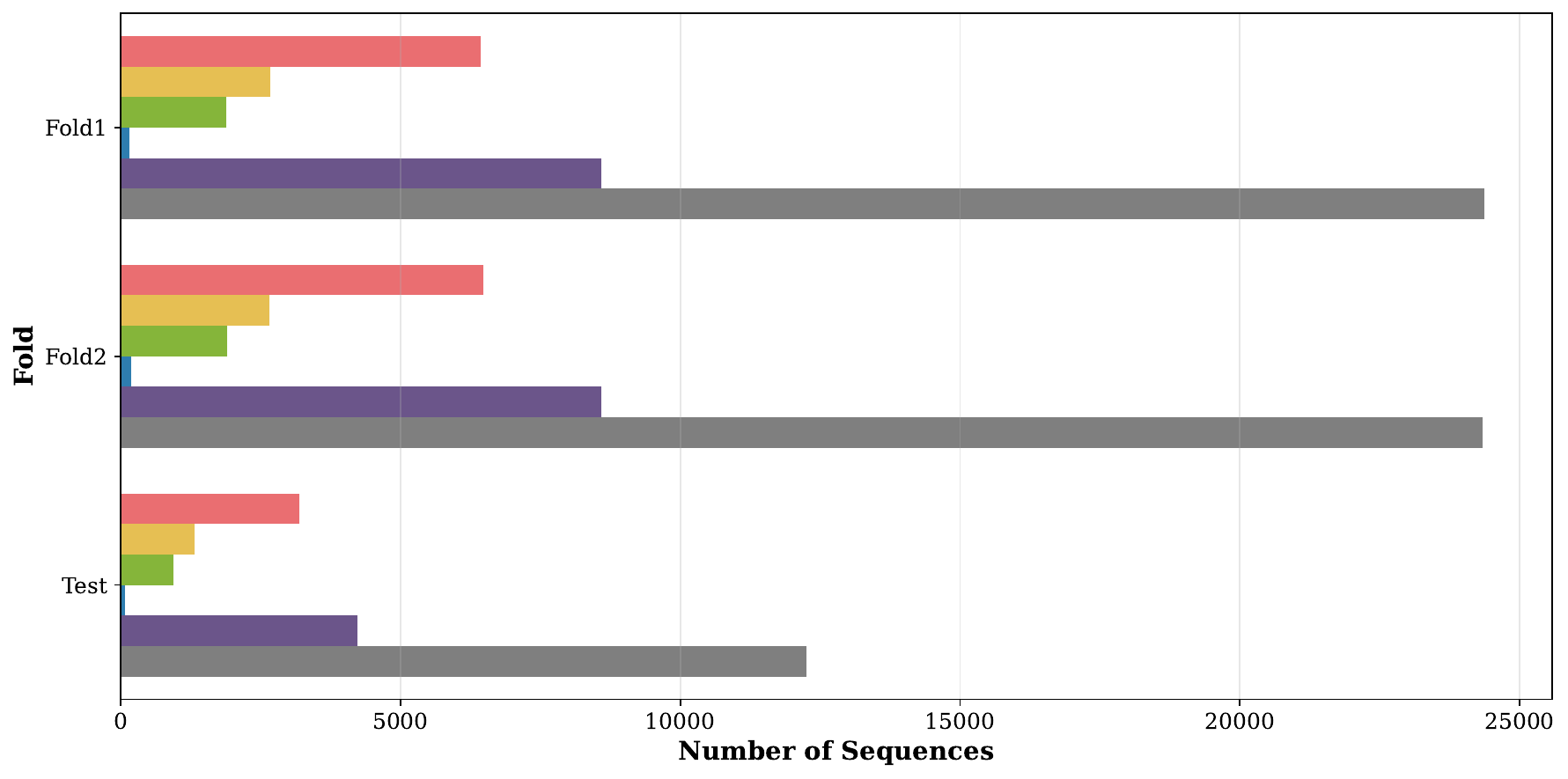}
      \caption{}
      \label{fig:fold-distribution}
    \end{subfigure}

    \vspace{0pt}
  \end{minipage}
    
    \caption{\textbf{Overview of \dataset{} Dataset Composition and Statistics.} (a) Multilabel distribution of AMPs across the four functional classes in \dataset{} Dataset, (b) Sequence length distribution for AMPs and non-AMPs, and (c) Distribution of AMP and non-AMP sequences in the two folds and the test set of the dataset.}
    \label{fig:dataset-composition}
\end{figure}

\subsection{Data Processing and Cleaning}

We remove sequences containing synthetic residues such as pyrrolysine (O), selenocysteine (U), $\beta$-alanine (Bal), 3-naphthylalanine (Nal), or 2-aminobutanoic acid (Abu), following the pipeline that AMP-Net \cite{ruiz2022rational} proposes. We exclude entries with undefined amino acids (X) and retain degenerate codes J, B, and Z, treating them as biologically valid representations of leucine/isoleucine, aspartic acid/asparagine, and glutamic acid/glutamine, respectively. We also enforce length constraints, retaining only peptides with lengths between 5 and 250 residues, ensuring structural relevance and alignment with established peptide standards \cite{ruiz2022rational}. Finally, to mitigate redundancy, we consolidate duplicate sequences across repositories and integrate their corresponding functional annotations into a unified multilabel vector. 

\subsection{ESCAPE Dataset} \label{subsection_escape_dataset}

The \dataset{} Dataset comprises \num{60950} non-AMPs and \num{21409} AMPs, which are functionally annotated into four major antimicrobial categories: antibacterial, antifungal, antiviral, and antiparasitic. Figure~\ref{fig:upset-plot} shows that most AMPs (\num{16106}) exhibit antibacterial activity, and \num{10924} belong exclusively to this class. The most common combination is antibacterial and antifungal sequences (\num{4960} peptides), while only \num{1671} peptides are solely antifungal. Antiviral entries total \num{4726} (\num{3479} unique), and antiparasitic peptides number \num{417} (\num{130} unique). Rarer multilabel groups include primarily antiparasitic sequences exhibiting antifungal or antiviral activity.

Fig. \ref{fig:violin-plot} presents the sequence length distribution for AMPs and non-AMPs. AMP sequences are predominantly centered around \num{30} amino acids, which is characteristic of this type of peptide structure \cite{Ma2024}. In contrast, non-AMPs exhibit a broader length range, with an average of \num{90} amino acids, and most sequences fall between \num{50} and \num{100} residues. This distinction highlights the inherent structural differences between AMPs and non-AMP sequences.

The \dataset{} Dataset is divided into three folds: two used for cross-validation and one reserved for independent testing. This partitioning strategy ensures comprehensive model evaluation by maintaining equal label distributions across folds, as shown in Fig. \ref{fig:fold-distribution}. This consistency supports reliable performance assessment and minimizes overfitting risks between functional classes.

\subsection{\dataset{} Benchmark} \label{subsection_benchmark}

To evaluate the performance of existing AMP classification methods under the standardized multilabel framework introduced by \dataset{}, we benchmark seven representative models with publicly available implementations that allow reproducibility: AMPlify \cite{li2022amplify}, AMP-BERT \cite{lee2023amp}, TransImbAMP \cite{pang2022integrating}, amPEPpy \cite{lawrence2021ampeppy}, AMPs-Net \cite{ruiz2022rational}, PEP-Net \cite{han2024pepnet}, and AVP-IFT \cite{avp2024}. We adapt each model to support multilabel classification and we train it with a two-fold cross-validation scheme. In Supplementary Section B, we provide per-model implementation details, with the training hyperparameters summarized in Supplementary Table \ref{tab:shared_hyperparameters}. To establish robust and consistent evaluation, we train each method three times with the random seeds (42, 1665, 8914) across all models. We perform the final evaluation with an ensemble that averages the probabilities from the two trained models. We assess performance using mean Average Precision (mAP) and F1-score, two widely used metrics for multilabel settings with severe class imbalance typical of AMP prediction \cite{puentes2025artificial}. We report the mean and standard deviation across the three runs.

\section{\dataset{} Baseline}\label{baseline_section}

Given the limitations of existing models in addressing multiclass classification of antimicrobial peptides, we design a method that combines sequential and structural information to perform the classification task on the ESCAPE Database. We introduce the ESCAPE Baseline model, a transformer-based architecture that integrates sequence and structural modalities through bidirectional cross-attention, and we provide a detailed account of its design and implementation. 

\subsection{Input Peptide Representation}

\textbf{Sequence Representation.} Let $\mathcal{S}$ denote the set of all peptide sequences. Each sequence $s \in \mathcal{S}$ is defined as an ordered list of amino acids $s = [a_1, a_2, \dots, a_{N}]$ with $N$ residues, where each $a_i \in \mathcal{A}$. The vocabulary $\mathcal{A}$ consists of \num{26} amino acid symbols (including rare and ambiguous codes), together with a special padding token, resulting in a vocabulary length of 27~\cite{li2022amplify}. Let \(\mathcal{T} = \{ t_1, t_2, \dots, t_{27} \}\) be a finite set of discrete tokens, with \(|\mathcal{T}| = 27\), representing the token vocabulary. We define a bijective mapping $f: \mathcal{A} \to \mathcal{T}$ such that for every $a_i \in \mathcal{A}$ there exists a unique token $t_i = f(a_i) \in \mathcal{T}$. This one-to-one correspondence allows each sequence $s$ to be equivalently represented as a token sequence $t = [t_1, t_2, \dots, t_{\mathcal{L}}]$ over the vocabulary $\mathcal{T}$. All sequences are either truncated or zero-padded to a fixed length $\mathcal{L}$.

\textbf{Structural Representation.} For each peptide, structural information is collected from the UniProt \cite{uniprot2019uniprot} and Protein Data Bank (PDB) \cite{bank1971protein} repositories. When experimental structures are unavailable, we use RosettaFold \cite{baek2021accurate} and AlphaFold3 \cite{abramson2024accurate} to predict the three-dimensional conformations. These models are state-of-the-art deep learning methods for protein structure prediction. Given a peptide with $N$ amino acids, we compute a distance matrix $\mathcal{M} \in \mathbb{R}^{N \times N}$, based on the 3D structure. Each element $\mathcal{M}_{i,j}$ corresponds to the Euclidean distance between the C$\alpha$ atoms of residues $i$ and $j$: $\mathcal{M}_{i,j} = \lVert \mathbf{r}_i - \mathbf{r}_j \rVert$, where $\mathbf{r}_{i,j} \in \mathbb{R}^3$ denotes the spatial coordinates of the $i^{\text{th}}$, $j^{\text{th}}$ residue. To ensure compatibility across peptides of varying lengths, $\mathcal{M}$ is resized to a fixed dimension of $224 \times 224$, enabling uniform input to the structural encoder.





\subsection{Model Architecture}\label{Architecture_subsection}

\begin{figure}[h!]
    \centering
    \includegraphics[width=\textwidth]{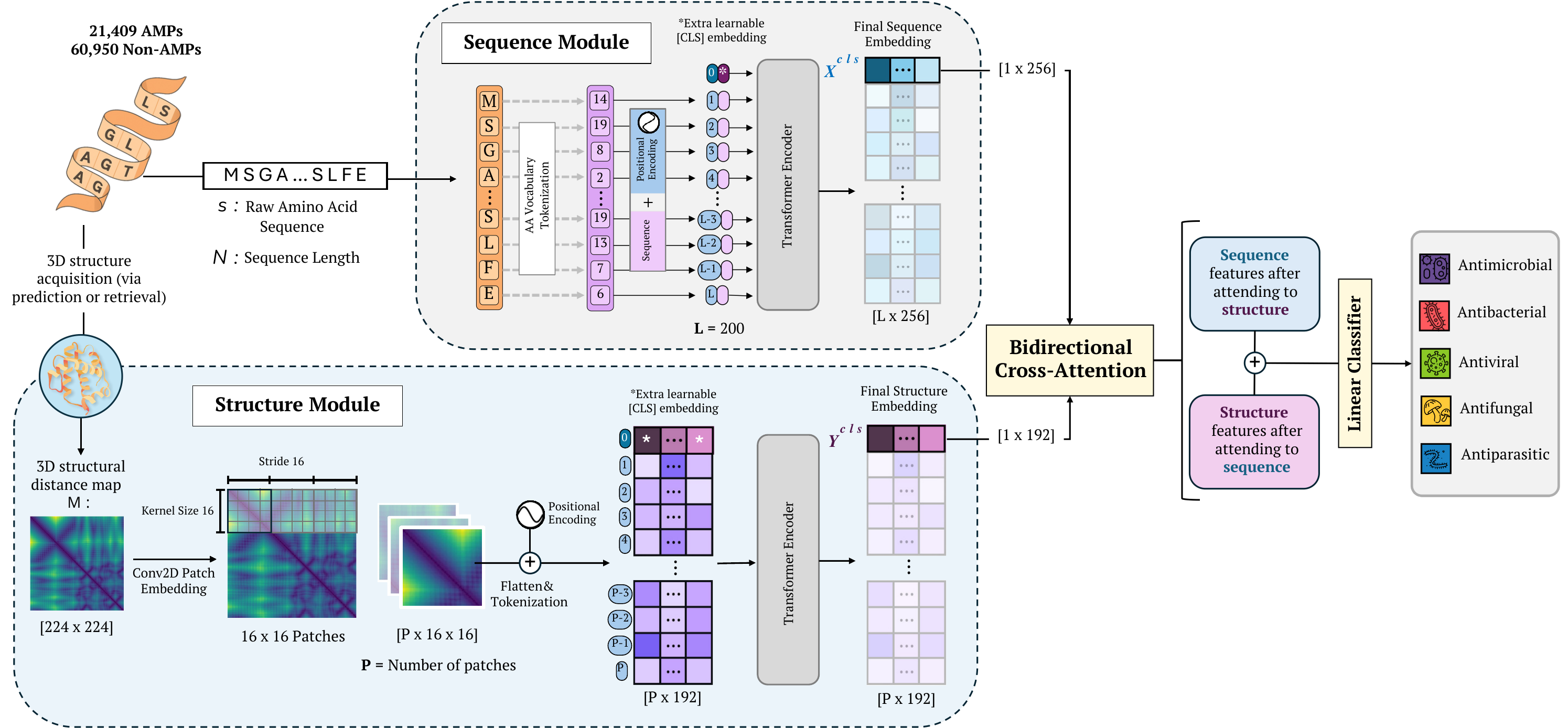}
    \caption{\textbf{\dataset{} Baseline Architecture Overview.} The model encodes each peptide using two parallel branches: the sequence module tokenizes amino acid residues. It extracts a [CLS] representation through a Transformer encoder. In contrast, the structure module processes a $224 \times 224$ distance matrix by embedding non-overlapping patches and applying a Transformer stack to produce a structural [CLS] token. A bidirectional cross-attention mechanism fuses these two representations by allowing each modality to attend to the other. The model concatenates the resulting attended CLS vectors and passes them through a linear layer to generate the final multilabel prediction vector.}

    \label{fig:overview-ampformer}
\end{figure}

The \dataset{} Baseline model is built upon a dual-branch transformer architecture designed to jointly encode the sequence and structural modalities of peptides, as illustrated in Fig.~\ref{fig:overview-ampformer}. Each branch independently processes one modality using a specific transformer encoder, and the resulting representations are fused via a bidirectional cross-attention mechanism to enable cross-domain interaction.

\textbf{Sequence Module.} The sequence branch takes the tokenized peptide sequence and maps each token to a 256-dimensional vector using a learnable embedding matrix. In addition, we include a special [CLS] token to capture global sequence-level information and add positional embeddings to preserve the order of amino acids. Later, this feeds the resulting embeddings through a stack of $\mathcal{D}$ Transformer encoder layers. This setup enables the model to learn local and long-range dependencies within the sequence. The [CLS] token at the output serves as a compact representation of the peptide’s primary structure.

\textbf{Structure Module.} The structural branch receives as input a single-channel $224 \times 224$ distance matrix $\mathcal{M}$, where each entry indicates the Euclidean distance between a pair of $C\alpha$ atoms in the peptide’s 3D structure. A 2D convolution with kernel size and stride of 16 partitions this matrix into non-overlapping $16 \times 16$ patches, producing a grid of flattened patches. The model projects each patch into a 192-dimensional embedding, creating a sequence of patch embeddings. The model adds a learnable [CLS] token at the beginning of the sequence to aggregate global structural information and appends fixed positional encodings to preserve spatial relationships. The model processes the sequence through a stack of $\mathcal{D}$ Transformer encoder layers that capture local and long-range spatial dependencies. The structure branch outputs the [CLS] token, which captures a compact representation of the peptide’s 3D conformation.

\textbf{Bidirectional Cross-Attention.} We apply a bidirectional cross-attention mechanism to integrate information from sequence and structure modalities after independently encoding each branch. As detailed in the Sequence and Structure Modules, our model processes the amino acid sequence and the distance matrix separately through their transformer encoders. Each encoder produces a contextual embedding matrix and a corresponding [CLS] token that summarizes global information.

Let $\mathbf{X} \in \mathbb{R}^{\mathcal{L} \times 256}$ and $\mathbf{Y} \in \mathbb{R}^{\mathcal{P} \times 192}$ denote the sequence and structure embedding matrices, respectively, where $\mathcal{L}$ and $\mathcal{P}$ are the modality-specific lengths, including the [CLS] token. The corresponding [CLS] embeddings are denoted as $\mathbf{X}_{\text{cls}}$ and $\mathbf{Y}_{\text{cls}}$.

To enable cross-modal interaction, we apply attention in both directions. First, the sequence attends to the structural features where $\mathbf{Q}_x = \mathbf{X} \mathbf{W}_Q^x$ are the queries derived from the sequence, and $\mathbf{K}_y = \mathbf{Y} \mathbf{W}_K^y$, $\mathbf{V}_y = \mathbf{Y} \mathbf{W}_V^y$ are the keys and values projected from the structure branch. The attention output $\mathbf{A}_x$ is added to the original sequence embeddings via a residual connection and refined through a feedforward network. Conversely, the structural representation attends to the sequence where $\mathbf{Q}_y = \mathbf{Y} \mathbf{W}_Q^y$, and $\mathbf{K}_x = \mathbf{X} \mathbf{W}_K^x$, $\mathbf{V}_x = \mathbf{X} \mathbf{W}_V^x$ are the projections from the sequence encoder.
\begin{align*}
\mathbf{A}_x &= \mathrm{softmax}\left( \frac{ \mathbf{Q}_x \mathbf{K}_y^\top }{ \sqrt{d} } \right) \mathbf{V}_y, \quad &
\mathbf{A}_y &= \mathrm{softmax}\left( \frac{ \mathbf{Q}_y \mathbf{K}_x^\top }{ \sqrt{d} } \right) \mathbf{V}_x,
\end{align*}

The bidirectional attention mechanism enables each modality to gather contextual information from the other by focusing on the most informative regions of the complementary representation. After bidirectional cross-attention, we concatenate the updated [CLS] from both modalities and pass the resulting vector through a classification head to produce the final prediction.

\subsection{Implementation Details}\label{Implementation_Details_Subsection}

We train the ESCAPE Baseline on a NVIDIA GPU A100 with a batch size \num{64} for \num{100} epochs, using the AdamW optimizer and a learning rate of $1 \times 10^{-4}$. We incorporate dropout layers within each transformer encoder block to mitigate overfitting and enhance generalization. In the sequence branch, we tokenize each input peptide and pad it to a fixed length of $\mathcal{S} = 200$, embedding each amino acid into a \num{256}-dimensional space. The sequence encoder comprises \num{4} transformer layers, each with \num{8} attention heads. In the structural branch, we represent each peptide’s distance matrix as a $224 \times 224$ single-channel image and divide it into non-overlapping $16 \times 16$ patches. We flatten each patch and project it into a \num{192}-dimensional embedding. We then process the resulting patch embeddings with a transformer encoder that shares the same configuration as the sequence branch, consisting of \num{4} layers and \num{8} attention heads, to capture spatial and geometric dependencies within the peptide structure.

\section{Experiments and Discussion} \label{Results}


\subsection{Main Results}\label{main_results}

Table~\ref{tab:f1-results} reports the mean and standard deviations over the three random seeds of the overall and per-class F1-scores, while Table~\ref{tab:main-results} presents the corresponding mAP values. The results demonstrate that our baseline consistently surpasses all seven state-of-the-art methods across both metrics, reaching relative improvements of \num{1.90}\% in overall F1-score and \num{2.56}\% in mAP compared to the second-best method.

The per-class evaluation reveals that ESCAPE Baseline achieves the most substantial gains in the least represented categories. In particular, it increases the AP for the antiparasitic class by \num{35.7}\% relative to the second-best method. Among sequence-based models, AMPlify attains the highest performance, with an average F1-score of \num{68.5}\% and mAP of \num{70.3}\%. AVP-IFT follows closely, achieving \num{66.5}\% F1 and \num{68.8}\% mAP by integrating physicochemical descriptors through a feature-augmented design. These comparisons indicate that no single modeling strategy, whether sequence-focused or feature-augmented, consistently dominates the others.

\input{Tables/f1_results}

Overall, performance trends highlight that the effectiveness of the model depends on the synergy between input representations and architectural design, rather than on the mere inclusion of additional descriptors. By combining sequence and structural modalities, ESCAPE Baseline leverages richer biological representations to improve generalization. The architecture also supports flexible operation under different configurations using only sequence information, only 3D structural data, or both, achieving its best performance when integrating the two. This versatility comes from its higher capacity and ability to align complementary modalities during training, establishing ESCAPE Baseline as both a robust and adaptable framework for multilabel AMP classification.

Furthermore, as shown in Figure \ref{fig:params} of the Supplementary Material, model size does not exhibit a consistent relationship with predictive performance across the evaluated methods. Notably, although the top-performing model has nearly 9 million parameters, the second-best model overall is based on a Random Forest classifier, making it the least computationally demanding method in our benchmark. Conversely, as Table~\ref{tab:f1-results} and Table~\ref{tab:main-results} report, BERT-based approaches do not rank among the top three performing models. These findings suggest that more complex architectures do not necessarily yield superior results for multilabel AMP classification. Additionally, our results underscore the limitations of large language models when applied to domains outside of natural language. Despite the domain adaptation efforts through BERT fine-tuning in models such as TransImbAMP and AMP-BERT, these methods fail to fully accommodate the peptide-specific data and, as a result, underperform in this task.


\input{Tables/ap_results}

Per-class analysis reveals substantial variation in predictive performance across functional
categories. As the number of samples per class decreases, most models exhibit a proportional decline in both mAP and F1-score, independent of their underlying architecture or feature design. The antiparasitic and antiviral classes, which contain the fewest examples, yield the lowest scores across all evaluated methods, highlighting the intrinsic difficulty of learning under severe data scarcity. In contrast, categories with broader representation, such as antibacterial and antifungal, display more stable results and narrower variability among models. This trend underscores the impact of label imbalance as the dominant factor shaping overall performance, suggesting that future improvements should focus on better representation learning rather than on increasing model complexity alone.

\subsection{Ablation Experiments}

To assess the individual contribution of each peptide representation modality in our baseline, we conduct an ablation experiment using the 42 seed. We evaluate three configurations: one using only the sequence module, another using only the structural module based on distance matrices, and a third combining both through the cross-attention module. Table~\ref{tab:main-ablations} shows that the sequence representation provides considerably more informative features than the structural one: with sequence-only input, our model achieves \num{21.7}\% higher mAP and \num{20.7}\% higher F1 score than with the distance matrix alone. This gap likely arises because the structural view captures spatial arrangement but omits explicit biochemical identities, thereby limiting the model’s ability to exploit residue-level patterns critical for antimicrobial activity. These findings highlight that the biological composition of the peptide encoded in the sequence plays a decisive role in classification performance. Yet Table~\ref{tab:main-ablations} also shows that combining both representations via cross-attention yields the best overall results: while the sequence-only variant is already strong, adding structural cues provides a complementary signal that further improves prediction quality.

\input{Tables/ablations}
As detailed in Section~\ref{baseline_section}, we obtain structural information for most peptides from UniProt and PDB, and infer the remaining structures using public generative models \cite{baek2021accurate, abramson2024accurate}. We also run a sensitivity experiment to assess the impact of using predicted structures. Supplementary Section C (Table \ref{tab:alphafold}) shows that relying solely on predicted structures leads to reduced performance compared to experimental structures, with absolute drops of \num{1.5}\% in mAP and \num{1.9}\% in F1. These results suggest that artificially generated structures may introduce additional sources of error inherent to those models, potentially degrading the quality of the structural data and impairing the model's ability to accurately classify peptides.


\subsection{Limitations and Broader Impact}\label{Limitations_Broader_Impact_Section}

In this work, we present a carefully curated and extensive dataset for AMP discovery. An important limitation arises from the scope of the domain, as the diversity of peptides in nature is vast and it is not feasible to capture all existing variants or ensure that our dataset fully represents the underlying distribution. Nonetheless, our benchmark provides a standardized and transparent framework for evaluating AMP classification models, helping to identify methodological gaps and guide future improvements. By promoting reproducibility and comparative analysis in this area, our work contributes to advancing computational tools for AMP discovery, which may support efforts in global health, particularly in the context of antibiotic resistance. However, further validation in real-world biological and clinical settings is required before deployment of such models.

Another limitation arises from the inherent differences in sequence length distributions. Antimicrobial peptides are naturally shorter than most non-AMPs, a characteristic tied to their biological function. In constructing ESCAPE, we aim to preserve this molecular distinction while avoiding strong correlations that could bias classification. The resulting dataset maintains realistic differences between classes without allowing sequence length to dominate predictive performance, supporting a fairer evaluation of computational models.

\subsection{Ethical Considerations}
From an ethical standpoint, ESCAPE is constructed entirely from publicly available, experimentally validated datasets, each distributed under its respective license. While the benchmark provides a foundation for advancing computational methods in antimicrobial peptide research, our contribution remains focused on methodological innovation rather than direct therapeutic or drug design applications.


At the same time, we acknowledge the potential risks arising from irresponsible use of this resource. Models trained on ESCAPE could, if misused, be employed to generate peptides without appropriate experimental validation, raising concerns about toxicity or biosecurity. To mitigate such risks, we encourage the research community to operate within established ethical, biosafety, and regulatory standards and to ensure that all experimental and computational findings are reported transparently and with accountability.


\section{Conclusions}

We introduce \textbf{\dataset{}}, the first standardized benchmark for multilabel antimicrobial peptide classification, designed to overcome key limitations of existing resources, including data fragmentation, inconsistent annotations, and limited functional scope. \dataset{} integrates over \num{80,000} peptides from \num{27} curated repositories into a biologically grounded multilabel framework encompassing antibacterial, antifungal, antiviral, antiparasitic and antimicrobial classes. It also includes a rigorously filtered set of non-antimicrobial sequences to support reliable supervised training and better reflect real-world prediction settings. Building upon this foundation, we propose a baseline using a transformer-based architecture that leverages sequence and structural information. We demonstrate that \dataset{} Benchmark enables fair and reproducible comparison across models and functional classes, setting a new standard for AI-driven AMP discovery, particularly in underrepresented categories such as antiviral and antiparasitic.

\section{Acknowledgements}

This research was partially funded by the Colombian Ministry of Science, Technology, and Innovation (Minciencias), under Cod. 1204-937-101846, CR 19576-2024 Call for Fundamental Research. This work was supported by Azure sponsorship credits granted by Microsoft’s AI for Good Research Lab.


\input{final_version.bbl}





\renewcommand\thesection{\Alph{section}} 
\renewcommand\thesubsection{\thesection.\arabic{subsection}} 
\renewcommand\thesubsubsection{\thesubsection.\arabic{subsubsection}}

\setcounter{section}{0}
\newpage

\begin{center}
\Large \textbf{Supplementary Material}\\[0.5em]
A Standardized Benchmark for Multilabel Antimicrobial Peptide Classification
\end{center}

\section{\dataset{} Dataset Compilation}

\subsection{Compilation and Standardization of Datasets}
  
 We compile ESCAPE from 27 peptide databases by systematically extracting experimentally validated antimicrobial peptides annotated for antibacterial, antifungal, antiparasitic, or antiviral activity. Databases exclusively focusing on a single category, such as AVPdb \cite{qureshi2014avpdb} (antiviral), are directly mapped to one of the four target classes. For negative examples, we filter peptides from external sources unrelated to antimicrobial activity, such as anticancer (e.g., CancerPPD \cite{tyagi2015cancerppd}, TumorHoPe\cite{kapoor2012tumorhope}) and neuroactive peptide databases (e.g., NeuroPep \cite{wang2015neuropep}, BrainPeps \cite{van2012brainpeps}). Additionally, we follow the methodology outlined in TransImbAMP\cite{pang2022integrating}, selecting non-antimicrobial peptides from UniProt \cite{uniprot2019uniprot} by applying strict exclusion criteria. Specifically, we discard sequences containing keywords such as “membrane,” “toxic,” “secretory,” “defensive,” “antibiotic,” “anticancer,” “antiviral,” or “antifungal” to enhance the quality of the negative class.
 
For large and hierarchically structured databases such as DBAASP\cite{pirtskhalava2021dbaasp}, DRAMP\cite{ma2025dramp}, dbAMP (with species-level annotations)\cite{yao2025dbamp}, and SATPdb (which lists 38 functional categories)\cite{singh2016satpdb}, we retain all peptides with annotations that map either directly or through hierarchical or taxonomic relationships to one of our four defined antimicrobial classes (antibacterial, antifungal, antiparasitic, antiviral). This includes entries annotated at the level of function (e.g., “antifungal”), target phenotype (e.g., “anti-Gram positive”), or biological taxonomy (e.g., species or genus) when these map to our label taxonomy. We exclude peptides whose annotations lack any functional or taxonomic correspondence to our classes. When databases separate targets by phenotype, such as “anti-Gram positive” and “anti-Gram negative,” we merge these into a unified antibacterial class. We also perform manual curation for complex hierarchical annotations, consolidating entries from species, family, or domain levels under the most appropriate class. After this initial selection, we identify and resolve duplicate sequences across datasets by merging complementary annotations. For instance, if a peptide appears in multiple sources with evidence of both antifungal and antibacterial activity, we retain a single entry enriched with both labels.

\subsection{Licenses and copyright}

To build the ESCAPE benchmark, we aggregate data from 27 publicly available peptide databases covering antibacterial, antifungal, antiparasitic, and antiviral peptides, as well as peptides with no known antimicrobial activity. Table~\ref{tab:scape-dataset-licenses} lists the number of peptides available in each source along with their corresponding license terms. We apply a rigorous filtering and selection pipeline to construct a legally compliant benchmark. For datasets under permissive licenses (e.g., CC BY or CC BY-NC), we include the relevant entries directly in ESCAPE. When licenses restrict redistribution (e.g., Oxford University Press\cite{novkovic2012dadp}\cite{kim2011neuropedia}\cite{qin2022dfbp}, Elsevier\cite{piotto2012yadamp}, Springer Nature\cite{van2012brainpeps}), we exclude the raw data and instead reference hashed identifiers and provide scripts to enable reproducibility. This strategy ensures that ESCAPE adheres to academic licensing standards while offering broad coverage of experimentally validated antimicrobial and non-antimicrobial peptides.

\begin{table}[htbp]
\caption{\textbf{Overview of peptide databases integrated into the ESCAPE benchmark}. Here we detail the number of peptides and associated licensing terms for each source.}
\centering
\renewcommand{\arraystretch}{1.3}
\begin{tabular}{>{\centering\arraybackslash}m{4.5cm}|
                >{\centering\arraybackslash}m{2.5cm}|
                >{\centering\arraybackslash}p{5cm}}
\midrule
\textbf{Database Name} & \textbf{Number of Peptides} & \textbf{License} \\
\midrule

BIOPEP-UWM Database \cite{minkiewicz2019biopep} & \num{3634} & \href{http://creativecommons.org/licenses/by/4.0/}{CC BY 4.0} \\
CPPsite 2.0 \cite{agrawal2016cppsite} & \num{1155} & \href{https://creativecommons.org/licenses/by-nc/4.0/}{CC BY-NC 4.0} \\
CAMPR3 \cite{waghu2016campr3} & \num{4519} & \href{https://creativecommons.org/licenses/by-nc/4.0/}{CC BY-NC 4.0} \\
TumorHoPe \cite{kapoor2012tumorhope} & \num{787} & \href{https://creativecommons.org/licenses/by/2.5/}{CC BY 2.5} \href{https://creativecommons.org/licenses/by/3.0/}{/3.0} \\
APD3\cite{wang2016apd3} & \num{3072} & \href{https://creativecommons.org/licenses/by-nc/4.0/}{CC BY-NC 4.0} \\
SPdb \cite{choo2005spdb}& \num{2512} & \href{https://creativecommons.org/licenses/by/2.0/}{CC BY 2.0} \\
ParaPep \cite{pretzel2013antiparasitic} & \num{194} & \href{https://creativecommons.org/licenses/by/4.0/}{CC BY \footnotemark[1]}\\
CancerPPD \cite{tyagi2015cancerppd} & \num{556} & \href{https://creativecommons.org/licenses/by-nc/4.0/}{CC BY-NC 4.0} \\
BrainPreps \cite{van2012brainpeps} & \num{92} & © Springer Nature \footnotemark[2] \\
Quorumpeps \cite{wynendaele2013quorumpeps} & \num{257} & \href{https://creativecommons.org/licenses/by-nc/3.0/}{CC BY-NC 3.0}\\
YADAMP \cite{piotto2012yadamp} & \num{2133} & © Elsevier \footnotemark[2] \\
LAMP2 \cite{zhao2013lamp} & \num{23253} & \href{https://creativecommons.org/licenses/by/4.0/}{CC BY \footnotemark[1]} \\
Milkampdb \cite{theolier2014milkamp} & \num{260} & \href{https://creativecommons.org/licenses/by/4.0/}{CC BY \footnotemark[1]} \\
DADP \cite{novkovic2012dadp} & \num{2557} & © Oxford University Press \footnotemark[2] \\
AntiTbPdb \cite{usmani2018antitbpdb} & \num{271} & \href{https://creativecommons.org/licenses/by/4.0/}{CC BY 4.0}\\
PeptideDB \cite{PeptideDB} & \num{1903} & \href{https://creativecommons.org/licenses/by/4.0/}{CC BY 4.0}\\
NeuroPrep \cite{wang2015neuropep} & \num{3875} & \href{https://creativecommons.org/licenses/by/4.0/}{CC BY 4.0} \\
SATPdb \cite{singh2016satpdb} & \num{9664} & \href{https://creativecommons.org/licenses/by-nc/4.0/}{CC BY-NC 4.0} \\
BioDADPep \cite{roy2019biodadpep} & \num{2543} & \href{https://creativecommons.org/licenses/by/4.0/}{CC BY \footnotemark[1]}\\
NeuroPedia \cite{kim2011neuropedia} & \num{847} & © Oxford University Press \footnotemark[2] \\
DFBP \cite{qin2022dfbp} & \num{7058} & © Oxford University Press \footnotemark[2]\\
dbAMP \cite{yao2025dbamp} & \num{35602} & \href{https://creativecommons.org/licenses/by-nc/4.0/}{CC BY-NC 4.0} \\
DRAMP \cite{ma2025dramp} & \num{11614} & \href{https://creativecommons.org/licenses/by/4.0/}{CC BY 4.0}\\
AVPdb \cite{qureshi2014avpdb} & \num{2683} &  \href{https://creativecommons.org/licenses/by/3.0/}{CC BY 3.0}\\
Hemolytik \cite{gautam2014hemolytik} & \num{523} & \href{https://creativecommons.org/licenses/by-nc/3.0/}{CC BY-NC 3.0} \\
DBAASP \cite{pirtskhalava2021dbaasp} & \num{22724} & \href{https://creativecommons.org/licenses/by-nc/4.0/}{CC BY-NC 4.0} \\
Uniprot \cite{uniprot2019uniprot} & \num{62453} &\href{https://creativecommons.org/licenses/by/4.0/}{CC BY 4.0}\\
\midrule
\end{tabular}
\label{tab:scape-dataset-licenses}
\end{table}

\footnotetext[1]{The dataset authors reference a general license (e.g., Creative Commons) without specifying the exact version or associated terms.}

\footnotetext[2]{The dataset authors do not explicitly state the license governing the use of their data, and reuse must follow the specific terms set by the respective publishers or journals under standard academic publishing policies. Attribution through proper citation of the original sources is required for any use of these datasets. To ensure compliance with these conditions, we configure our dataset access pipeline to retrieve data directly through the official APIs or download interfaces provided by the original sources.}

\subsection{Statistical Analysis of the \dataset{} Dataset}

\begin{figure}[t]
    \centering
    \begin{subfigure}[b]{0.9\textwidth}
        \centering
        \includegraphics[width=\textwidth, keepaspectratio]{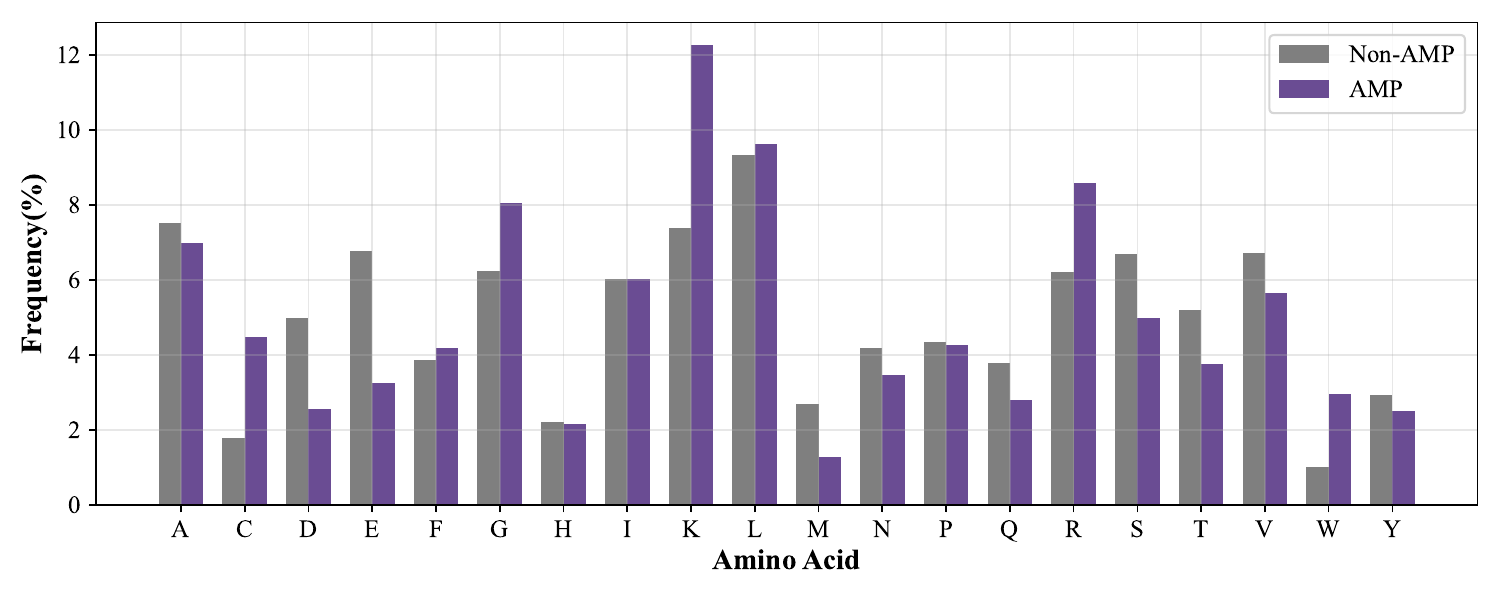}
        \caption{}
        \label{fig:comp_dist}
    \end{subfigure}

    \vspace{0.5cm} 

    \begin{subfigure}[b]{0.9\textwidth}
        \centering
        \includegraphics[width=0.97\textwidth, keepaspectratio]{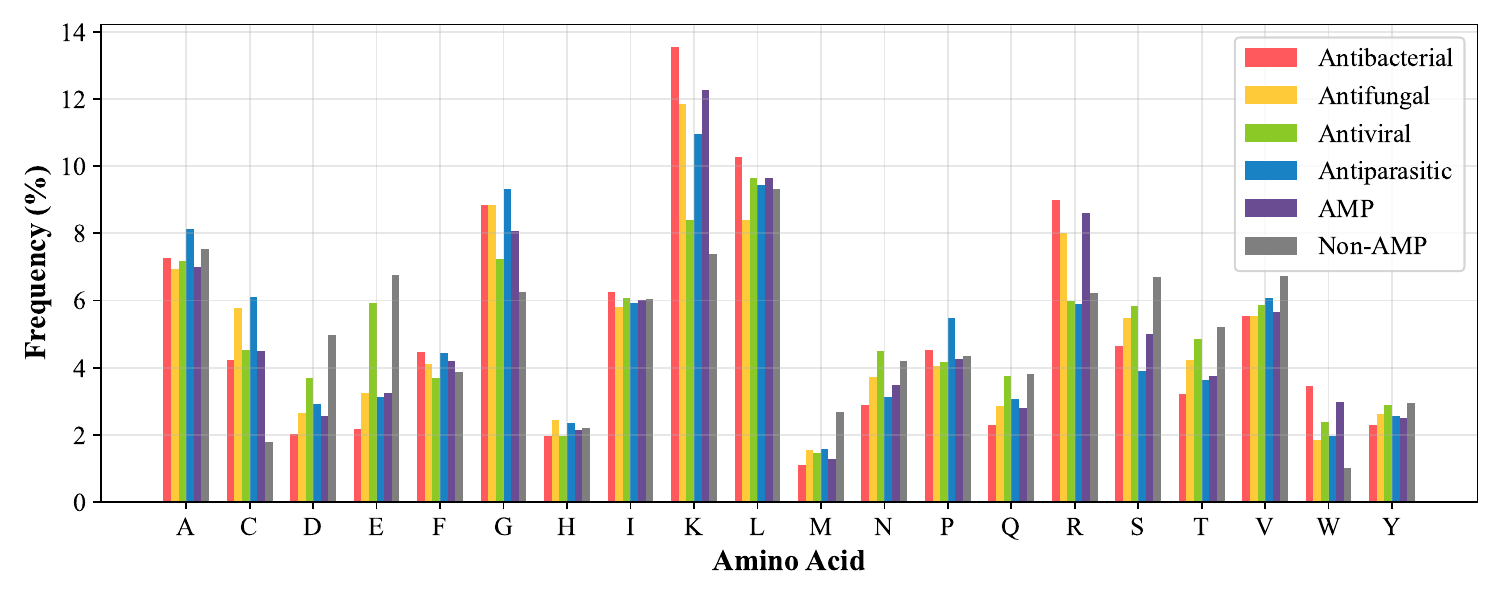}
        \caption{}
        \label{fig:normalized_dist}
    \end{subfigure}
    
    \caption{\textbf{Comparison of amino acid distributions in the \dataset{} dataset.} (a) Amino acid distributions for AMPs and Non-AMPs, with frequency differences reflecting variations between functional and non-functional peptides. (b) Normalized amino acid distributions with respect to each class for the multilabel clasification task. Overall, the dataset maintains a consistent aminoacid composition across categories.}
    \label{fig:vertical_subplots}
\end{figure}

We analyze the amino acid distribution across the five classes in the \dataset{} Database: antibacterial, antifungal, antiviral, antiparasitic, and antimicrobial peptides (AMP), along with the non antimicrobial peptides (Non-AMP). The frequencies vary across categories, reflecting differences in peptide counts. As an example, we compare the amino acid distributions between AMP and Non-AMP in Figure~\ref{fig:comp_dist}. Since the Non-AMP category contains \num{2.85} times more peptides than AMP, its amino acid frequencies are higher. Figure~\ref{fig:normalized_dist} shows the normalized amino acid distributions across \dataset{} dataset classes and Non-AMP. Despite differences in the number of peptides per category, the relative frequencies remain largely consistent. This suggests that the dataset maintains a coherent overall amino acid composition across categories, with minimal variation. Moreover, it indicates that the underlying sequence composition remains stable, even across functionally distinct peptide groups.

\section{AMP Models on the \dataset{} Benchmark}
\label{sup:Supplementary_ESCAPE_Benchmark}

\subsection{Implementation Details}

We implement and evaluate all models mentioned in Section 3.4 to address the multilabel classification task. To ensure a comprehensive and representative benchmark, the evaluation includes a diverse set of model architectures: an attention-based LSTM \cite{li2022amplify}, a random forest classifier \cite{lawrence2021ampeppy}, a graph neural network \cite{ruiz2022rational}, and four Transformer-based models \cite{lee2023amp, pang2022integrating, han2024pepnet, avp2024}, two of which leverage BERT backbones \cite{lee2023amp, pang2022integrating} and two that employ the vanilla Transformer architecture with physicochemical and sequence-related features \cite{han2024pepnet, avp2024}. Moreover, AVP-IFT \cite{avp2024} also employs a contrastive learning module. Table~\ref{tab:shared_hyperparameters} summarizes the training hyperparameters used for all models evaluated in this study, with the exception of amPEPpy \cite{lawrence2021ampeppy}. For amPEPpy, which is based on a random forest classifier, we employ an ensemble of \num{160} bootstrap-aggregated decision trees and assess generalization performance using out-of-the-bag (OOB) estimation.

To address the multilabel classification task, we configure each model to output a binary vector of length five, where each dimension corresponds to one of the target antimicrobial classes: antibacterial, antiviral, antifungal, antiparasitic, and antimicrobial. We apply a sigmoid activation function to the final layer to produce independent probability estimates for each class. To adapt AVP-IFT \cite{avp2024} to the multilabel classification task, we changed the original binary similarity label in the contrastive loss to a continuous value that represents the fraction of similarity and disimilarity across the five classes in the multilabel vector. We train each model separately on two distinct data folds and use a cross-validation setup to encourage generalization and reduce overfitting. During inference, we average the output logits from both trained instances before applying the sigmoid activation. This ensembling strategy treats both models as equal contributors and integrates their predictions into a single output.

\input{Tables/hyperparameters}


\subsection{Statistical Significance of AMP Models in the \dataset{} Benchmark}

\begin{figure}[t!]
    \centering
    \includegraphics[width=0.8\textwidth]{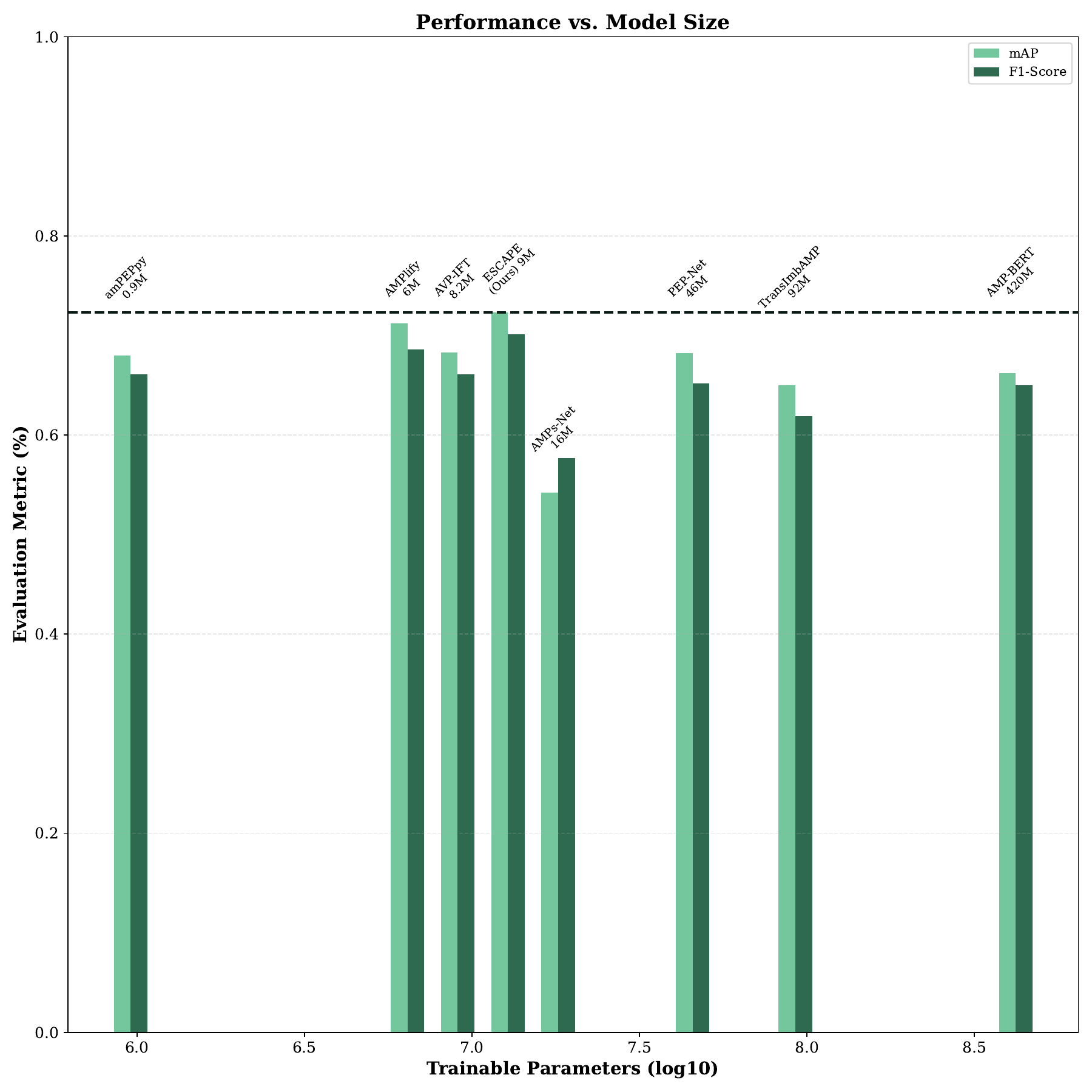}
    \caption{\textbf{Comparison of model performance and number of trainable parameters across all evaluated methods.} Since lighter models like the ESCAPE Baseline and AMPlify \cite{li2022amplify} show the best ensemble results in the test split and heavier models (e.g., BERT-based transformers \cite{pang2022integrating} \cite{lee2023amp}) yield lower performance, we observe no consistent correlation between model size and predictive capability. Specifically, the ESCAPE Baseline achieves the best overall results with a fraction of the parameters used by large transformer models, suggesting that performance gains can be attained without increased model complexity. }
    \label{fig:params}
\end{figure}

We adopt a two-fold cross-validation strategy by training each model independently on two complementary folds of the dataset. For each trained instance, we evaluate performance on the corresponding test set using overall metrics, namely mean average precision (mAP) and F1 score, as well as class-wise scores for the five antimicrobial categories. We report these metrics separately for each fold to assess statistical consistency and variability across data partitions. To summarize model performance, we calculate the mean and standard deviation across the two folds, providing a reliable estimate of average predictive accuracy and performance variability. For this evaluation, we report the results with the 42 seed. Table~\ref{tab:f1_sup_results} presents the F1-scores, while Table~\ref{tab:AP_sup_results} reports the corresponding mAP values.

\input{Tables/f1_Folds_Supplementary}
In Section 5.1, we report the performance of the ESCAPE baseline, defined as the ensemble of two independently trained models with logits averaged prior to the sigmoid activation. The ESCAPE Baseline ensemble using seed 42 achieves an overall F1 score of \num{69.5} and a mAP of \num{72.7}, surpassing the mean performance of individual folds by \num{3.35} and \num{5.90} points, respectively. These results highlight the benefit of combining complementary representations learned from distinct training partitions, leading to improved robustness and predictive accuracy.

\input{Tables/ap_Folds_Supplementary}

We analyze the relationship between model performance and the number of trainable parameters to examine the trade-off between predictive accuracy and architectural complexity. Figure~\ref{fig:params} presents this comparison across all evaluated methods. Models with fewer parameters frequently achieve higher mAP and F1 scores, indicating that increased model size does not inherently translate to improved predictive capacity. However, when focusing on the amPEPpy, AMPlify, and ESCAPE models, a more consistent trend emerges: performance gains align with substantial increases in model size. For instance, achieving an improvement of less than 10\% in mAP and F1 requires scaling from 0.9 million to over 6 million parameters. These findings underscore the importance of balancing model complexity with practical performance benefits when designing architectures for antimicrobial peptide classification.

\section{Sensitivity of the model to predicted 3D protein structures}\label{Sensitivity_3D_Structures}

To further evaluate the influence of structural inputs on the ESCAPE Baseline, we conducted an ablation study in which we replaced experimental 3D crystal structures with AlphaFold-predicted counterparts. Experimental structures were available for only 2.086 peptides from UniProt (2.5\% of ESCAPE), distributed as 846 in Fold 1, 825 in Fold 2, and 415 in Test. In our experiment, the experimental structures in Fold 1 and Fold 2 (1.671 peptides) were replaced with their AlphaFold predictions, while the Test set remained unchanged. Results indicate that using only predicted structures reduces performance relative to experimental data (Table \ref{tab:alphafold}), with absolute drops of 1.5\% (mAP) and 1.9\% (F1). These findings confirm that experimental crystal structures provide superior inputs for the structure module, but also show that predicted structures remain a viable alternative when experimental data are unavailable. This sensitivity highlights the dependence of the ESCAPE Baseline on the quality of structural representations, suggesting that future improvements in structure prediction methods, such as AlphaFold \cite{abramson2024accurate} and RosettaFold \cite{baek2021accurate}, may directly enhance classification performance.

\begin{table}[h!]
\centering
\caption{\textbf{Ablation experiments for the ESCAPE Baseline with respect to 3D predicted protein structures.} These results correspond to the 42 random seed trained model.}
\begin{tabular}{c|c|c}
\toprule
\textbf{Training Data} & \textbf{mAP} & \textbf{F1-Score} \\
\midrule
Only generated structures           & 71.2 & 67.5 \\
Experimental + generated structures & 72.7 & 69.4 \\
\bottomrule
\end{tabular}
\label{tab:alphafold}
\end{table}

\newpage
\section*{NeurIPS Paper Checklist}

\begin{enumerate}

\item {\bf Claims}
    \item[] Question: Do the main claims made in the abstract and introduction accurately reflect the paper's contributions and scope?
    \item[] Answer: \answerYes{} 
    \item[] Justification: The abstract and introduction accurately articulate the dataset and benchmark proposed in Section \ref{Escape_Dataset_section} and then the baseline model proposed in Section \ref{baseline_section}.
    \item[] Guidelines:
    \begin{itemize}
        \item The answer NA means that the abstract and introduction do not include the claims made in the paper.
        \item The abstract and/or introduction should clearly state the claims made, including the contributions made in the paper and important assumptions and limitations. A No or NA answer to this question will not be perceived well by the reviewers. 
        \item The claims made should match theoretical and experimental results, and reflect how much the results can be expected to generalize to other settings. 
        \item It is fine to include aspirational goals as motivation as long as it is clear that these goals are not attained by the paper. 
    \end{itemize}

\item {\bf Limitations}
    \item[] Question: Does the paper discuss the limitations of the work performed by the authors?
    \item[] Answer: \answerYes{} 
    \item[] Justification: Yes, we address the limitations of our work in Section \ref{Limitations_Broader_Impact_Section}.

    \item[] Guidelines:
    \begin{itemize}
        \item The answer NA means that the paper has no limitation while the answer No means that the paper has limitations, but those are not discussed in the paper. 
        \item The authors are encouraged to create a separate "Limitations" section in their paper.
        \item The paper should point out any strong assumptions and how robust the results are to violations of these assumptions (e.g., independence assumptions, noiseless settings, model well-specification, asymptotic approximations only holding locally). The authors should reflect on how these assumptions might be violated in practice and what the implications would be.
        \item The authors should reflect on the scope of the claims made, e.g., if the approach was only tested on a few datasets or with a few runs. In general, empirical results often depend on implicit assumptions, which should be articulated.
        \item The authors should reflect on the factors that influence the performance of the approach. For example, a facial recognition algorithm may perform poorly when image resolution is low or images are taken in low lighting. Or a speech-to-text system might not be used reliably to provide closed captions for online lectures because it fails to handle technical jargon.
        \item The authors should discuss the computational efficiency of the proposed algorithms and how they scale with dataset size.
        \item If applicable, the authors should discuss possible limitations of their approach to address problems of privacy and fairness.
        \item While the authors might fear that complete honesty about limitations might be used by reviewers as grounds for rejection, a worse outcome might be that reviewers discover limitations that aren't acknowledged in the paper. The authors should use their best judgment and recognize that individual actions in favor of transparency play an important role in developing norms that preserve the integrity of the community. Reviewers will be specifically instructed to not penalize honesty concerning limitations.
    \end{itemize}

\item {\bf Theory assumptions and proofs}
    \item[] Question: For each theoretical result, does the paper provide the full set of assumptions and a complete (and correct) proof?
    \item[] Answer: \answerNA{} 
    \item[] Justification: The paper does not include theoretical results that require formal proofs.
    \item[] Guidelines:
    \begin{itemize}
        \item The answer NA means that the paper does not include theoretical results. 
        \item All the theorems, formulas, and proofs in the paper should be numbered and cross-referenced.
        \item All assumptions should be clearly stated or referenced in the statement of any theorems.
        \item The proofs can either appear in the main paper or the supplemental material, but if they appear in the supplemental material, the authors are encouraged to provide a short proof sketch to provide intuition. 
        \item Inversely, any informal proof provided in the core of the paper should be complemented by formal proofs provided in appendix or supplemental material.
        \item Theorems and Lemmas that the proof relies upon should be properly referenced. 
    \end{itemize}

    \item {\bf Experimental result reproducibility}
    \item[] Question: Does the paper fully disclose all the information needed to reproduce the main experimental results of the paper to the extent that it affects the main claims and/or conclusions of the paper (regardless of whether the code and data are provided or not)?
    \item[] Answer: \answerYes{} 
    \item[] Justification: Section \ref{subsection_escape_dataset} details the procedure to create the proposed dataset with clearly defined partitions that strictly follow the experimental protocol. For complete reproducibility, the Supplementary Material provides full details. Section \ref{baseline_section} describes the training procedure and the information needed to reproduce baseline results.
    
    \item[] Guidelines:
    \begin{itemize}
        \item The answer NA means that the paper does not include experiments.
        \item If the paper includes experiments, a No answer to this question will not be perceived well by the reviewers: Making the paper reproducible is important, regardless of whether the code and data are provided or not.
        \item If the contribution is a dataset and/or model, the authors should describe the steps taken to make their results reproducible or verifiable. 
        \item Depending on the contribution, reproducibility can be accomplished in various ways. For example, if the contribution is a novel architecture, describing the architecture fully might suffice, or if the contribution is a specific model and empirical evaluation, it may be necessary to either make it possible for others to replicate the model with the same dataset, or provide access to the model. In general. releasing code and data is often one good way to accomplish this, but reproducibility can also be provided via detailed instructions for how to replicate the results, access to a hosted model (e.g., in the case of a large language model), releasing of a model checkpoint, or other means that are appropriate to the research performed.
        \item While NeurIPS does not require releasing code, the conference does require all submissions to provide some reasonable avenue for reproducibility, which may depend on the nature of the contribution. For example
        \begin{enumerate}
            \item If the contribution is primarily a new algorithm, the paper should make it clear how to reproduce that algorithm.
            \item If the contribution is primarily a new model architecture, the paper should describe the architecture clearly and fully.
            \item If the contribution is a new model (e.g., a large language model), then there should either be a way to access this model for reproducing the results or a way to reproduce the model (e.g., with an open-source dataset or instructions for how to construct the dataset).
            \item We recognize that reproducibility may be tricky in some cases, in which case authors are welcome to describe the particular way they provide for reproducibility. In the case of closed-source models, it may be that access to the model is limited in some way (e.g., to registered users), but it should be possible for other researchers to have some path to reproducing or verifying the results.
        \end{enumerate}
    \end{itemize}

\item {\bf Open access to data and code}
    \item[] Question: Does the paper provide open access to the data and code, with sufficient instructions to faithfully reproduce the main experimental results, as described in supplemental material?
    \item[] Answer: \answerYes{} 
    \item[] Justification:The \dataset{} Dataset is already available \href{https://doi.org/10.7910/DVN/C69MCD}{here}, and the source code for \dataset{} Baseline is already available \href{https://github.com/BCV-Uniandes/ESCAPE}{here}.
    \item[] Guidelines:
    \begin{itemize}
        \item The answer NA means that paper does not include experiments requiring code.
        \item Please see the NeurIPS code and data submission guidelines (\url{https://nips.cc/public/guides/CodeSubmissionPolicy}) for more details.
        \item While we encourage the release of code and data, we understand that this might not be possible, so “No” is an acceptable answer. Papers cannot be rejected simply for not including code, unless this is central to the contribution (e.g., for a new open-source benchmark).
        \item The instructions should contain the exact command and environment needed to run to reproduce the results. See the NeurIPS code and data submission guidelines (\url{https://nips.cc/public/guides/CodeSubmissionPolicy}) for more details.
        \item The authors should provide instructions on data access and preparation, including how to access the raw data, preprocessed data, intermediate data, and generated data, etc.
        \item The authors should provide scripts to reproduce all experimental results for the new proposed method and baselines. If only a subset of experiments are reproducible, they should state which ones are omitted from the script and why.
        \item At submission time, to preserve anonymity, the authors should release anonymized versions (if applicable).
        \item Providing as much information as possible in supplemental material (appended to the paper) is recommended, but including URLs to data and code is permitted.
    \end{itemize}

\item {\bf Experimental setting/details}
    \item[] Question: Does the paper specify all the training and test details (e.g., data splits, hyperparameters, how they were chosen, type of optimizer, etc.) necessary to understand the results?
    \item[] Answer: \answerYes{} 
    \item[] Justification: Section \ref{Implementation_Details_Subsection} details the hyperparameters and type of optimizer for our model baseline. We include these details for the other models in the Supplementary Material. Section \ref{subsection_benchmark} describes the data splits used for all models. These sections ensure both reproducibility and interpretability of the reported results.
    \item[] Guidelines:
    \begin{itemize}
        \item The answer NA means that the paper does not include experiments.
        \item The experimental setting should be presented in the core of the paper to a level of detail that is necessary to appreciate the results and make sense of them.
        \item The full details can be provided either with the code, in appendix, or as supplemental material.
    \end{itemize}

\item {\bf Experiment statistical significance}
    \item[] Question: Does the paper report error bars suitably and correctly defined or other appropriate information about the statistical significance of the experiments?
    \item[] Answer: \answerYes{} 
    \item[] Justification: The Supplementary Material provides the results with error bars for each fold, ensuring a complete understanding of the statistical significance. The main results in Section \ref{main_results} include error bars as we report the ensemble from the 2 cross-folds for each model in the benchmark across 3 random seeds.
    \item[] Guidelines:
    \begin{itemize}
        \item The answer NA means that the paper does not include experiments.
        \item The authors should answer "Yes" if the results are accompanied by error bars, confidence intervals, or statistical significance tests, at least for the experiments that support the main claims of the paper.
        \item The factors of variability that the error bars are capturing should be clearly stated (for example, train/test split, initialization, random drawing of some parameter, or overall run with given experimental conditions).
        \item The method for calculating the error bars should be explained (closed form formula, call to a library function, bootstrap, etc.)
        \item The assumptions made should be given (e.g., Normally distributed errors).
        \item It should be clear whether the error bar is the standard deviation or the standard error of the mean.
        \item It is OK to report 1-sigma error bars, but one should state it. The authors should preferably report a 2-sigma error bar than state that they have a 96\% CI, if the hypothesis of Normality of errors is not verified.
        \item For asymmetric distributions, the authors should be careful not to show in tables or figures symmetric error bars that would yield results that are out of range (e.g. negative error rates).
        \item If error bars are reported in tables or plots, The authors should explain in the text how they were calculated and reference the corresponding figures or tables in the text.
    \end{itemize}

\item {\bf Experiments compute resources}
    \item[] Question: For each experiment, does the paper provide sufficient information on the computer resources (type of compute workers, memory, time of execution) needed to reproduce the experiments?
    \item[] Answer: \answerYes{} 
    \item[] Justification: The computational resources used for the experiments are detailed in Section \ref{Implementation_Details_Subsection}.
    
    \item[] Guidelines:
    \begin{itemize}
        \item The answer NA means that the paper does not include experiments.
        \item The paper should indicate the type of compute workers CPU or GPU, internal cluster, or cloud provider, including relevant memory and storage.
        \item The paper should provide the amount of compute required for each of the individual experimental runs as well as estimate the total compute. 
        \item The paper should disclose whether the full research project required more compute than the experiments reported in the paper (e.g., preliminary or failed experiments that didn't make it into the paper). 
    \end{itemize}
    
\item {\bf Code of ethics}
    \item[] Question: Does the research conducted in the paper conform, in every respect, with the NeurIPS Code of Ethics \url{https://neurips.cc/public/EthicsGuidelines}?
    \item[] Answer: \answerYes{} 
    \item[] Justification: ESCAPE uses publicly available data, follows community adopted standards for dataset curation, and does not involve human subjects or sensitive personal information. All contributions comply with the NeurIPS Code of Ethics.

    \item[] Guidelines:
    \begin{itemize}
        \item The answer NA means that the authors have not reviewed the NeurIPS Code of Ethics.
        \item If the authors answer No, they should explain the special circumstances that require a deviation from the Code of Ethics.
        \item The authors should make sure to preserve anonymity (e.g., if there is a special consideration due to laws or regulations in their jurisdiction).
    \end{itemize}

\item {\bf Broader impacts}
    \item[] Question: Does the paper discuss both potential positive societal impacts and negative societal impacts of the work performed?
    \item[] Answer: \answerYes{} 
    \item[] Justification: We address potential societal impacts in Section \ref{Limitations_Broader_Impact_Section}. The paper highlights positive impacts by advancing antimicrobial peptide discovery to support public health efforts against antibiotic resistance and notes the need for further validation. However, all results derived from this benchmark require subsequent experimental validation before any practical or societal application, substantially reducing the risk of negative societal impacts.
    
    \item[] Guidelines:
    \begin{itemize}
        \item The answer NA means that there is no societal impact of the work performed.
        \item If the authors answer NA or No, they should explain why their work has no societal impact or why the paper does not address societal impact.
        \item Examples of negative societal impacts include potential malicious or unintended uses (e.g., disinformation, generating fake profiles, surveillance), fairness considerations (e.g., deployment of technologies that could make decisions that unfairly impact specific groups), privacy considerations, and security considerations.
        \item The conference expects that many papers will be foundational research and not tied to particular applications, let alone deployments. However, if there is a direct path to any negative applications, the authors should point it out. For example, it is legitimate to point out that an improvement in the quality of generative models could be used to generate deepfakes for disinformation. On the other hand, it is not needed to point out that a generic algorithm for optimizing neural networks could enable people to train models that generate Deepfakes faster.
        \item The authors should consider possible harms that could arise when the technology is being used as intended and functioning correctly, harms that could arise when the technology is being used as intended but gives incorrect results, and harms following from (intentional or unintentional) misuse of the technology.
        \item If there are negative societal impacts, the authors could also discuss possible mitigation strategies (e.g., gated release of models, providing defenses in addition to attacks, mechanisms for monitoring misuse, mechanisms to monitor how a system learns from feedback over time, improving the efficiency and accessibility of ML).
    \end{itemize}
    
\item {\bf Safeguards}
    \item[] Question: Does the paper describe safeguards that have been put in place for responsible release of data or models that have a high risk for misuse (e.g., pretrained language models, image generators, or scraped datasets)?
    \item[] Answer: \answerNA{} 
    \item[] Justification: The paper does not involve models or data with a high risk of misuse. The dataset consists of curated peptide sequences from licensed public sources, and we cite properly all datasets and comply with its usage terms.

    \item[] Guidelines:
    \begin{itemize}
        \item The answer NA means that the paper poses no such risks.
        \item Released models that have a high risk for misuse or dual-use should be released with necessary safeguards to allow for controlled use of the model, for example by requiring that users adhere to usage guidelines or restrictions to access the model or implementing safety filters. 
        \item Datasets that have been scraped from the Internet could pose safety risks. The authors should describe how they avoided releasing unsafe images.
        \item We recognize that providing effective safeguards is challenging, and many papers do not require this, but we encourage authors to take this into account and make a best faith effort.
    \end{itemize}

\item {\bf Licenses for existing assets}
    \item[] Question: Are the creators or original owners of assets (e.g., code, data, models), used in the paper, properly credited and are the license and terms of use explicitly mentioned and properly respected?
    \item[] Answer: \answerYes{} 
    \item[] Justification: We properly cite all employed methods from other authors and respect the licenses of the datasets that contributed to the creation of the main dataset. The Supplementary Material provides more information about these licenses.
    \item[] Guidelines:
    \begin{itemize}
        \item The answer NA means that the paper does not use existing assets.
        \item The authors should cite the original paper that produced the code package or dataset.
        \item The authors should state which version of the asset is used and, if possible, include a URL.
        \item The name of the license (e.g., CC-BY 4.0) should be included for each asset.
        \item For scraped data from a particular source (e.g., website), the copyright and terms of service of that source should be provided.
        \item If assets are released, the license, copyright information, and terms of use in the package should be provided. For popular datasets, \url{paperswithcode.com/datasets} has curated licenses for some datasets. Their licensing guide can help determine the license of a dataset.
        \item For existing datasets that are re-packaged, both the original license and the license of the derived asset (if it has changed) should be provided.
        \item If this information is not available online, the authors are encouraged to reach out to the asset's creators.
    \end{itemize}

\item {\bf New assets}
    \item[] Question: Are new assets introduced in the paper well documented and is the documentation provided alongside the assets?
    \item[] Answer: \answerNA{} 
    \item[] Justification: The paper does not include new assets. Section \ref{data_compilation} shows all the datasets from which this benchmark was compiled.
    
    \item[] Guidelines:
    \begin{itemize}
        \item The answer NA means that the paper does not release new assets.
        \item Researchers should communicate the details of the dataset/code/model as part of their submissions via structured templates. This includes details about training, license, limitations, etc. 
        \item The paper should discuss whether and how consent was obtained from people whose asset is used.
        \item At submission time, remember to anonymize your assets (if applicable). You can either create an anonymized URL or include an anonymized zip file.
    \end{itemize}

\item {\bf Crowdsourcing and research with human subjects}
    \item[] Question: For crowdsourcing experiments and research with human subjects, does the paper include the full text of instructions given to participants and screenshots, if applicable, as well as details about compensation (if any)? 
    \item[] Answer: \answerNA{}{} 
    \item[] Justification: The paper does not involve any experiments with human subjects or crowdsourcing.

    \item[] Guidelines:
    \begin{itemize}
        \item The answer NA means that the paper does not involve crowdsourcing nor research with human subjects.
        \item Including this information in the supplemental material is fine, but if the main contribution of the paper involves human subjects, then as much detail as possible should be included in the main paper. 
        \item According to the NeurIPS Code of Ethics, workers involved in data collection, curation, or other labor should be paid at least the minimum wage in the country of the data collector. 
    \end{itemize}

\item {\bf Institutional review board (IRB) approvals or equivalent for research with human subjects}
    \item[] Question: Does the paper describe potential risks incurred by study participants, whether such risks were disclosed to the subjects, and whether Institutional Review Board (IRB) approvals (or an equivalent approval/review based on the requirements of your country or institution) were obtained?
    \item[] Answer: \answerNA{} 
    \item[] Justification: The paper does not involve research with human subjects.
    \item[] Guidelines:
    \begin{itemize}
        \item The answer NA means that the paper does not involve crowdsourcing nor research with human subjects.
        \item Depending on the country in which research is conducted, IRB approval (or equivalent) may be required for any human subjects research. If you obtained IRB approval, you should clearly state this in the paper. 
        \item We recognize that the procedures for this may vary significantly between institutions and locations, and we expect authors to adhere to the NeurIPS Code of Ethics and the guidelines for their institution. 
        \item For initial submissions, do not include any information that would break anonymity (if applicable), such as the institution conducting the review.
    \end{itemize}

\item {\bf Declaration of LLM usage}
    \item[] Question: Does the paper describe the usage of LLMs if it is an important, original, or non-standard component of the core methods in this research? Note that if the LLM is used only for writing, editing, or formatting purposes and does not impact the core methodology, scientific rigorousness, or originality of the research, declaration is not required.
    \item[] Answer: \answerNA{} 
    \item[] Justification: This research does not involve LLMs as a core component.
    \item[] Guidelines: 
    \begin{itemize}
        \item The answer NA means that the core method development in this research does not involve LLMs as any important, original, or non-standard components.
        \item Please refer to our LLM policy (\url{https://neurips.cc/Conferences/2025/LLM}) for what should or should not be described.
    \end{itemize}

\end{enumerate}

\end{document}

%% file: Tables/f1_results.tex
\begin{table}[h!]
\centering
\caption{\textbf{Overall and Per-Class F1-Scores on the ESCAPE Benchmark.} F1-scores for each model averaged over the 42, 1665, and 8914 random seeds on the 5-class multilabel classification task in the ESCAPE Benchmark (\%).}
\label{tab:f1-results}
\resizebox{\textwidth}{!}{
\begin{tabular}{c|c|ccccc}
\toprule
\textbf{Method} & \textbf{F1-Score} & \textbf{Antibacterial} & \textbf{Antiviral} & \textbf{Antifungal} & \textbf{Antiparasitic} & \textbf{Antimicrobial} \\
\midrule
AMPs-Net \cite{ruiz2022rational} & $57.7\pm0.70$ & $78.9\pm0.77$ & $59.2\pm0.79$ & $61.1\pm0.51$ & $5.9\pm0.71$  & $83.5\pm0.79$ \\
TranslmbAMP \cite{pang2022integrating} & $62.0\pm0.70$ & $87.1\pm0.96$ & $59.2\pm0.50$ & $54.7\pm0.51$ & $21.8\pm0.81$ & $87.2\pm0.75$ \\
AMP-BERT \cite{lee2023amp} & $64.7\pm0.64$ & $89.3\pm0.27$ & $63.0\pm0.95$ & $60.2\pm0.26$ & $20.6\pm3.52$ & $90.5\pm0.22$ \\
PEP-Net \cite{han2024pepnet} & $65.5\pm0.61$ & $\mathbf{89.5\pm0.10}$ & $58.1\pm0.78$ & $\mathbf{65.2\pm0.55}$ & $22.8\pm0.61$ & $\mathbf{91.2\pm0.15}$ \\
amPEPpy \cite{lawrence2021ampeppy} & $66.5\pm0.37$ & $87.6\pm0.07$ & $61.6\pm2.02$ & $60.4\pm1.90$ & $34.7\pm0.98$ & $90.9\pm3.78$ \\
AVP-IFT \cite{avp2024} & $66.5\pm0.59$ & $89.1\pm0.47$ & $\mathbf{64.8\pm0.06}$ & $60.7\pm0.55$ & $28.0\pm3.84$ & $89.9\pm0.31$ \\
AMPlify \cite{li2022amplify} & $68.5\pm0.77$ & $88.8\pm0.26$ & $60.0\pm1.05$ & $65.0\pm1.57$ & $40.9\pm2.48$ & $90.0\pm0.30$ \\
\textbf{ESCAPE Baseline (Ours)} & $\mathbf{69.8\pm0.43}$ & $88.8\pm0.34$ & $64.4\pm0.88$ & $61.0\pm0.75$ & $\mathbf{44.8\pm0.50}$ & $90.0\pm0.32$ \\
\bottomrule
\end{tabular}
}
\end{table}

%% file: Tables/ap_results.tex
\begin{table}[h!]
\centering
\caption{\textbf{Mean and Per-Class AP Results on the ESCAPE Benchmark.} AP for each model averaged over the 42, 1665, and 8914 random seeds on the 5-class multilabel classification task in the ESCAPE Benchmark (\%).}
\label{tab:main-results}
\resizebox{\textwidth}{!}{
\begin{tabular}{c|c|ccccc} 
\toprule
\textbf{Method} & \textbf{mAP} & \textbf{Antibacterial} & \textbf{Antiviral} & \textbf{Antifungal} & \textbf{Antiparasitic} & \textbf{Antimicrobial} \\ 
\midrule

AMPs-Net \cite{ruiz2022rational} & $54.6\pm0.86$ & $82.5\pm0.72$ & $51.2\pm0.88$ & $53.1\pm0.84$ & $5.3\pm0.67$ & $82.1\pm0.80$ \\

TransImbAMP \cite{pang2022integrating} & $64.9\pm1.11$ & $92.5\pm1.23$ & $65.0\pm1.63$ & $56.3\pm0.96$ & $16.7\pm0.86$ & $94.0\pm0.90$ \\

AMP-BERT \cite{lee2023amp} & $66.9\pm1.17$ & $92.3\pm0.59$ & $65.9\pm1.84$ & $61.5\pm2.28$ & $21.4\pm2.61$ & $93.6\pm1.25$ \\

amPEPpy \cite{lawrence2021ampeppy} & $68.5\pm0.48$ & $93.9\pm0.24$ & $67.7\pm0.28$ & $62.2\pm0.27$ & $23.8\pm1.61$ & $95.2\pm0.05$ \\

PEP-Net \cite{han2024pepnet} & $68.4\pm0.53$ & \bm{$95.2\pm0.21$} & $61.2\pm0.67$ & $\bm{72.6\pm0.78}$ & $16.2\pm0.84$ & $\bm{96.7\pm0.26}$ \\

AVP-IFT \cite{avp2024} & $68.8\pm0.50$ & $94.3\pm0.49$ & $\bm{71.1\pm0.36}$ & $63.3\pm1.36$ & $20.0\pm4.25$ & $95.5\pm0.50$ \\

AMPlify \cite{li2022amplify} & $70.3\pm0.87$ & $94.0\pm0.19$ & $66.1\pm5.56$ & $68.3\pm4.27$ & $27.7\pm1.33$ & $95.3\pm0.31$ \\

\textbf{ESCAPE Baseline (Ours)} & $\bm{72.1\pm0.60}$ & $94.2\pm0.21$ & $69.8\pm0.46$ & $63.4\pm0.74$ & $\bm{37.6\pm2.87}$ & $95.6\pm0.04$ \\

\bottomrule
\end{tabular}}
\end{table}

%% file: Tables/ablations.tex
\begin{table}[h!]
\centering
\caption{\textbf{ESCAPE Baseline Ablation Experiments.} mAP (\%) and Overall F1-score (\%) reported for the 42 random seed trained model.}
\label{tab:main-ablations}
\resizebox{.75\textwidth}{!}{
\begin{tabular}{c|c|c|c||c|c} 
\toprule
\textbf{Structure Module} & 
\textbf{Sequence Module} & 
\textbf{Cross Attention} & 
\textbf{mAP} & 
\textbf{F1} \\ 
\midrule
\checkmark  & $\cdot$ & $\cdot$ & 47.7 & 46.9 \\
$\cdot$ & \checkmark & $\cdot$ & 69.4 & 67.6 \\
\checkmark & \checkmark & \checkmark & \textbf{72.7} & \textbf{69.5} \\

\bottomrule
\end{tabular}}
\end{table}

%% file: Tables/hyperparameters.tex
\begin{table}[h!]
\caption{\textbf{Training hyperparameters for the implemented AMP deep learning models.} For each deep learning model trained in the \dataset{} Benchmark we show the training and model architecture hyperparameters.}
\label{tab:shared_hyperparameters}
\centering
\resizebox{\textwidth}{!}{%
\begin{tabular}{
  >{\centering\arraybackslash}p{2.3cm} | 
  >{\centering\arraybackslash}p{1.4cm} 
  >{\centering\arraybackslash}p{1.4cm} 
  >{\centering\arraybackslash}p{1.8cm} 
  >{\centering\arraybackslash}p{1.4cm} 
  >{\centering\arraybackslash}p{1.4cm}
  >{\centering\arraybackslash}p{1.6cm}
}
\toprule
\textbf{Hyperparameter} & \textbf{AMPlify} \cite{li2022amplify} & \textbf{AMP-BERT} \cite{lee2023amp} & \textbf{TranslmbAMP} \cite{pang2022integrating} & \textbf{AMPs-Net} \cite{ruiz2022rational} & \textbf{PEP-Net} \cite{han2024pepnet} & \textbf{AVP-IFT} \cite{avp2024} \\
\midrule
Max Length & 200 & 200 & 180 & $-$ & 40 & 250 \\
Batch Size & 32 & 1 & 64 & 64 & 256 & 64 \\
Epochs & 70 & 15 & 256 & 300 & 100 & 100 \\
Learning Rate & $1 \cdot 10^{-3}$ & $1 \cdot 10^{-5}$ & $4 \cdot 10^{-2}$ & $5 \cdot 10^{-5}$ & $1 \cdot 10^{-4}$ & $1 \cdot 10^{-3}$ \\
Optimizer & Adam & Adam & Adam & Adamax & Adam & Adam \\
Dropout Rate & 0.5/0.2 & 0.0 & 0.2 & 0.2 & 0.5 & 0.5 \\
Hidden Dimension & 512 & 1024 & 512 & 256 & 1024 & 566 \\
Attention Heads & 32 & 16 & 12 & $-$ & 4 & 2 \\
Activation & ReLU & GELU & Leaky ReLU & ReLU & ELU & ReLU \\
Transformer Layers & $-$ & 30 & 12 & $-$ & 1 & 1 \\
\bottomrule
\end{tabular}
} 
\end{table}

%% file: Tables/f1_Folds_Supplementary.tex
\begin{table}[h!]
\centering
\caption{\textbf{Overall and Per-Class F1-Scores on the ESCAPE Benchmark without the logits ensemble strategy.} F1-scores for each model on the 5-class multilabel classification task in the ESCAPE Benchmark (\%). We report values as mean ± standard deviation. These results correspond to the 42 random seed trained models.}
\label{tab:f1_sup_results}
\resizebox{\textwidth}{!}{
\begin{tabular}{c|c|ccccc}
\toprule
\textbf{Method} & \textbf{F1-Score} & \textbf{Antibacterial} & \textbf{Antiviral} & \textbf{Antifungal} & \textbf{Antiparasitic} & \textbf{Antimicrobial} \\
\midrule
AMPs-Net \cite{ruiz2022rational} & $55.75 \pm 0.78$ & $77.9 \pm 0.57$ & $56.3 \pm 1.41$ & $53.65 \pm 0.77$ & $4.0 \pm 0.14$ & $80.65 \pm 1.48$ \\

TranslmbAMP \cite{pang2022integrating} & $60.21 \pm 0.68$ & $86.03 \pm 0.03$ & $56.6 \pm 0.87$ & $52.05 \pm 0.76$ & $20.36 \pm 1.65$ & $85.99 \pm 0.13$ \\

AMP-BERT \cite{lee2023amp} & $63.49 \pm 0.54$ & $86.99 \pm 0.15$ & $58.56 \pm 0.21$ & $54.07 \pm 1.24$ & $20.37 \pm 2.31$ & $87.75 \pm 0.54$ \\

PEP-Net \cite{han2024pepnet} & $62.86 \pm 1.89$ & $\bm{88.52 \pm 0.27}$ & $\bm{60.93 \pm 3.84}$ & $55.55 \pm 2.04$ & $19.46 \pm 2.63$ & $\bm{89.80 \pm 0.65}$ \\

amPEPpy \cite{lawrence2021ampeppy} & $63.60 \pm 0.17$ & $86.71 \pm 0.14$ & $59.32 \pm 0.63$ & $55.61 \pm 1.75$ & $28.30 \pm 1.50$ & $88.08 \pm 0.17$ \\

AVP-IFT \cite{avp2024} & $61.32 \pm 0.60$ & $87.27 \pm 0.02$ & $59.83 \pm 0.21$ & $55.42 \pm 1.34$ & $15.74 \pm 3.71$ & $88.32 \pm 0.44$ \\

AMPlify \cite{li2022amplify} & $65.10 \pm 0.78$ & $87.61 \pm 0.14$ & $59.32 \pm 0.63$ & $55.55 \pm 1.37$ & $32.95 \pm 1.70$ & $88.04 \pm 0.52$ \\

\textbf{ESCAPE Baseline (Ours)} & $\bm{66.15 \pm 0.07}$ & $86.85 \pm 0.35$ & $58.88 \pm 1.06$ & $\bm{56.00 \pm 0.00}$ & $\bm{40.45 \pm 2.19}$ & $88.55 \pm 0.21$ \\

\bottomrule
\end{tabular}
}
\end{table}

%% file: Tables/ap_Folds_Supplementary.tex
\begin{table}[h!]
\centering
\caption{\textbf{Mean and Per-Class AP Results on the ESCAPE Benchmark without the logits ensemble strategy.} AP for each model on the 5-class multilabel classification task in the ESCAPE Benchmark (\%). We report values as mean ± standard deviation. These results correspond to the 42 random seed trained models.}
\label{tab:AP_sup_results}
\resizebox{\textwidth}{!}{
\begin{tabular}{c|c|ccccc} 
\toprule
\textbf{Method} & \textbf{mAP} & \textbf{Antibacterial} & \textbf{Antiviral} & \textbf{Antifungal} & \textbf{Antiparasitic} & \textbf{Antimicrobial} \\ 
\midrule

AMPs-Net \cite{ruiz2022rational} & $51.50 \pm 0.85$ & $80.30 \pm 1.27$ & $47.30 \pm 1.98$ & $49.50 \pm 1.77$ & $4.70 \pm 0.14$ & $79.15 \pm 0.92$ \\

TransImbAMP \cite{pang2022integrating} & $62.57 \pm 0.11$ & $91.98 \pm 0.08$ & $62.21 \pm 0.25$ & $52.48 \pm 0.01$ & $13.27 \pm 0.56$ & $92.90 \pm 0.14$ \\

AMP-BERT \cite{lee2023amp} & $64.44 \pm 0.36$ & $90.58 \pm 0.65$ & $63.52 \pm 0.05$ & $56.35 \pm 0.52$ & $16.91 \pm 0.13$ & $91.41 \pm 0.06$ \\

amPEPpy \cite{lawrence2021ampeppy} & $65.12 \pm 0.17$ & $93.30 \pm 0.15$ & $63.24 \pm 0.68$ & $\bm{57.83} \bm{\pm} \bm{0.39}$ & $16.38 \pm 0.32$ & $94.84 \pm 0.07$ \\

PEP-Net \cite{han2024pepnet} & $65.43 \pm 1.58$ & $\bm{94.54} \bm{\pm} \bm{0.17}$ & $\bm{67.79} \bm{\pm} \bm{4.44}$ & $57.61 \pm 1.53$ & $12.14 \pm 2.23$ & $\bm{95.48} \bm{\pm} \bm{0.16}$ \\

AVP-IFT \cite{avp2024} & $62.08 \pm 1.13$ & $91.30 \pm 1.98$ & $63.50 \pm 1.79$ & $55.78 \pm 1.28$ & $7.38 \pm 2.04$ & $92.46 \pm 1.13$ \\

AMPlify \cite{li2022amplify} & $64.25 \pm 0.01$ & $91.14 \pm 1.57$ & $61.02 \pm 0.86$ & $55.01 \pm 0.04$ & $21.55 \pm 4.29$ & $92.52 \pm 1.84$ \\

\textbf{ESCAPE Baseline (Ours)} & $\bm{66.80} \bm{\pm} \bm{0.42}$ & $91.55 \pm 0.35$ & $63.25 \pm 0.35$ & $56.70 \pm 1.56$ & $\bm{28.50} \bm{\pm} \bm{0.42}$ & $94.05 \pm 0.07$ \\

\bottomrule
\end{tabular}}
\end{table}